\documentclass{article}

\usepackage{microtype}
\usepackage{graphicx}
\usepackage{subfigure}
\usepackage{booktabs} 

\usepackage{algorithm}
\usepackage{algorithmic}

\usepackage{newfloat}
\usepackage{listings}
\usepackage{pifont}   
\usepackage{xcolor}

\usepackage[accepted]{icml2026}

\usepackage{amsmath}
\usepackage{amssymb}
\usepackage{mathtools}
\usepackage{amsthm}
\usepackage{caption} 
\usepackage{colortbl}
\usepackage{adjustbox}
\usepackage{graphicx}
\usepackage{booktabs}
\usepackage{caption}
\usepackage{multirow}
\usepackage{amsmath}
\usepackage{amssymb}
\usepackage{helvet}
\usepackage{longtable}
\usepackage{wrapfig}
\usepackage{bbding}

\floatstyle{ruled}
\newfloat{listing}{tb}{lst}{}
\floatname{listing}{Listing}

\usepackage{hyperref}
\hypersetup{
    colorlinks=true,
    linkcolor=blue!40!black,
    citecolor=blue!40!black,
    urlcolor=blue!40!black
}

\usepackage[capitalize,noabbrev]{cleveref}

\theoremstyle{plain}

\theoremstyle{definition}

\theoremstyle{remark}

\usepackage[textsize=tiny]{todonotes}

\begin{document}

\twocolumn[

\icmltitle{Diffusion-Guided Pretraining for Brain Graph Foundation Models}



\icmlsetsymbol{equal}{*}

\begin{icmlauthorlist}
\icmlauthor{Xinxu Wei}{1}
\icmlauthor{Rong Zhou}{2}
\icmlauthor{Lifang He}{2}
\icmlauthor{Yu Zhang}{3}

\end{icmlauthorlist}

\icmlaffiliation{1}{Department of Electrical and Computer Engineering, Lehigh University, Bethlehem, PA, USA}
\icmlaffiliation{2}{Department of Computer Science and Engineering, Lehigh University, Bethlehem, PA, USA}
\icmlaffiliation{3}{Department of Psychiatry and Behavioral Sciences, Stanford University School of Medicine, Stanford, CA, USA}

\icmlcorrespondingauthor{Yu Zhang}{yzhangsu@stanford.edu}
\icmlcorrespondingauthor{Lifang He}{lih319@lehigh.edu}

\icmlkeywords{Machine Learning, ICML}

\vskip 0.3in
]

\printAffiliationsAndNotice{}




\begin{abstract}

With the growing interest in foundation models for brain signals, graph-based pretraining has emerged as a promising paradigm for learning transferable representations from connectome data. However, existing contrastive and masked autoencoder methods typically rely on naive random dropping or masking for augmentation, which is ill-suited for brain graphs and hypergraphs as it disrupts semantically meaningful connectivity patterns. Moreover, commonly used graph-level readout and reconstruction schemes fail to capture global structural information, limiting the robustness of learned representations. In this work, we propose a unified diffusion-based pretraining framework that addresses both limitations. First, diffusion is designed to guide structure-aware dropping and masking strategies, preserving brain graph semantics while maintaining effective pretraining diversity. Second, diffusion enables topology-aware graph-level readout and node-level global reconstruction by allowing graph embeddings and masked nodes to aggregate information from globally related regions. Extensive experiments across multiple neuroimaging datasets with over 25,000 subjects and 60,000 scans involving various mental disorders and brain atlases demonstrate consistent performance improvements.

\end{abstract}

\section{Introduction}

\begin{figure}[htbp]
\centering
	\includegraphics[width=8.2cm]{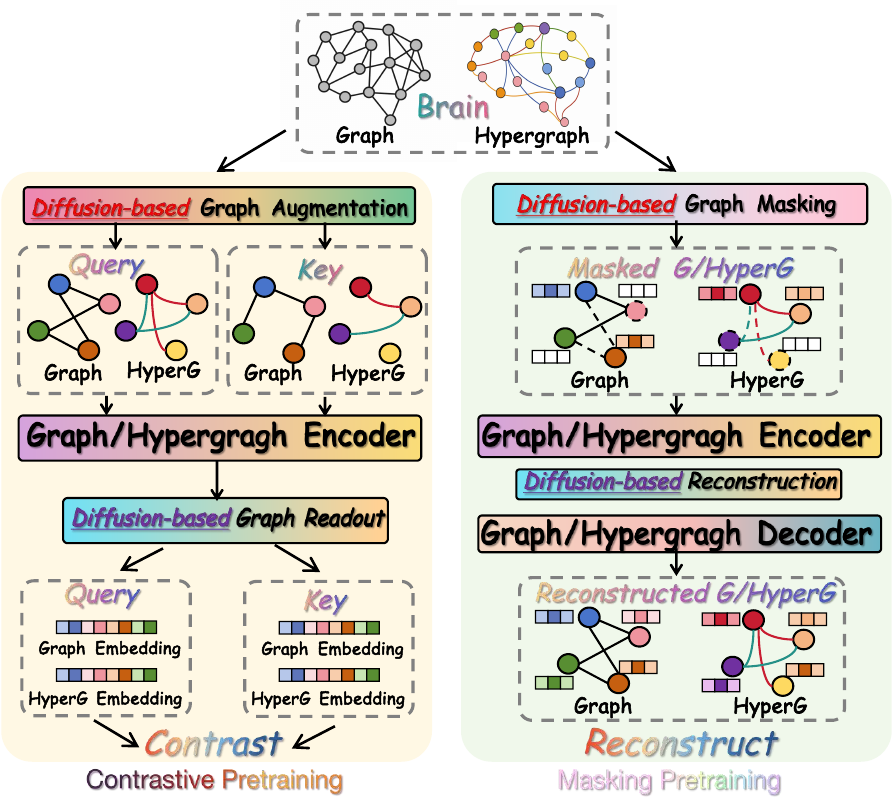}
    \vspace{-20pt}
	\caption{The proposed diffusion-enhanced framework includes Contrastive Pretraining (left) and Masking Pretraining (right) for Graph/Hypergraph. In both pipelines, graph diffusion is applied for topology-aware augmentation, masking, diffusion-based readout and reconstruction.}
	\label{model}
\vspace{-10pt}
\end{figure}

Recent advances in large language models \citep{achiam2023gpt, touvron2023llama} have demonstrated the effectiveness of foundation models \citep{liu2023towards} pretrained on large-scale unlabeled data. Inspired by this success, the neuroscience community has increasingly explored foundation models for brain signals \citep{caro2023brainlm, wei2025brain, yang2024brainmass, dong2024brain}, aiming to learn transferable representations that generalize across subjects, brain atlases, and neurological conditions.
Most existing brain foundation models are pretrained directly on raw time series \citep{caro2023brainlm} or connectome data \citep{yang2024brainmass}, however, graph-based foundation models \citep{liu2023towards} offer a more effective and efficient alternative by explicitly modeling brain topology while avoiding the heavy computational cost of sequence-level pretraining.
Brain connectivity data \citep{bullmore2011brain}, derived from neuroimaging modalities such as fMRI \citep{smith2013resting} and EEG \citep{wei2025multi}, are naturally represented as graphs \citep{kipf2016semi} or hypergraphs \citep{feng2019hypergraph}, where nodes correspond to brain regions and edges or hyperedges encode pairwise or higher-order interactions. Pretraining graph- and hypergraph-based models on connectome data therefore provides a promising pathway toward scalable and reusable brain representation learning.

Despite recent progress, existing graph pretraining paradigms are not well aligned with the structural characteristics of brain graphs and hypergraphs.
Current graph pretraining methods are mainly built upon two representative paradigms: graph contrastive learning (GCL) \citep{you2020graph, qiu2020gcc} and graph masked autoencoder (GMAE) \citep{hou2022graphmae, hou2023graphmae2}.
While these approaches have shown strong performance on generic graph benchmarks, directly applying them to brain graphs and hypergraphs introduces several fundamental limitations.
First, both GCL and GMAE typically rely on naive random dropping or masking of nodes, edges, or hyperedges as their core augmentation or corruption mechanisms.
Such perturbations are ill-suited for brain graphs, where global connectivity patterns encode critical functional semantics.
If important nodes or edges are excessively dropped or masked, the semantic integrity of the brain graph may be severely distorted; conversely, if perturbations predominantly affect less informative components, the resulting training signals become weak and uninformative.
As a result, random augmentation and masking can easily produce unrealistic contrastive views or unstable reconstruction targets, leading to fragile representations \citep{ding2022data, zhao2022graph}.
Second, existing pretraining frameworks largely overlook global structural aggregation.
In GCL, graph-level representations are usually obtained via structure-agnostic readout functions that fail to integrate long-range dependencies across the graph.
Similarly, in GMAE, masked nodes are commonly reconstructed using only local neighborhood information, without leveraging global context from distant but structurally related regions.
This locality bias further limits the expressiveness and robustness of learned representations for brain graphs and hypergraphs, where long-range interactions are essential.
These limitations highlight the need for more principled graph pretraining strategies that are better aligned with the structural and semantic properties of brain graphs and hypergraphs, and are essential for building effective brain graph foundation models.

To address these challenges, as shown in Figure \ref{model}, we propose a unified diffusion-based pretraining framework for brain graphs and hypergraphs.
Our core insight is that graph diffusion provides a principled way to encode global structural relationships \citep{gasteiger2019diffusion}, which can be directly embedded into the key operations of graph contrastive learning and graph masked autoencoder pretraining.
For graph contrastive learning pretraining, diffusion guides structure-aware graph augmentation, where nodes, edges, or hyperedges are dropped according to their global structural influence rather than uniformly at random.
This preserves semantically important brain connectivity patterns while maintaining sufficient diversity for contrastive learning.
Diffusion is further incorporated into the graph-level readout, enabling node representations to aggregate information from structurally related regions before pooling, resulting in topology-aware and robust graph embeddings.
For graph masked autoencoder pretraining, diffusion is used to guide masking and enable global reconstruction.
Diffusion-guided masking avoids excessive corruption of critical components, while diffusion-based reconstruction allows masked nodes or connections to leverage information from distant but structurally related regions, alleviating the locality bias of conventional masked autoencoder methods.

Our main contributions are summarized as follows:
\vspace{-5pt}
\begin{itemize}
    \item We propose a unified diffusion-based pretraining framework that integrates graph diffusion into both graph contrastive learning and graph masked autoencoder pretraining for brain graphs and hypergraphs.
    \item We leverage diffusion to design brain-aware graph augmentation strategies, enabling structure-guided dropping and masking that safely preserve semantic connectivity patterns while maintaining sufficient diversity.
    \item We exploit diffusion for topology-aware graph-level readout and node-level global reconstruction, allowing graph representations and masked components to aggregate information from structurally related regions.
\end{itemize}

\section{Related Works}

\subsection{Brain Foundation Models}

Recent work has explored foundation models for brain signals, inspired by large-scale pretraining in language and vision. Time-series-based approaches such as BrainLM \citep{caro2023brainlm} and Brain-JEPA \citep{dong2024brain} pretrain neural representations directly from brain signals using masked modeling. Beyond the temporal domain, connectome-based models such as BrainMass \citep{yang2024brainmass} leverage functional connectivity matrices for pretraining, while BrainGFM \citep{wei2025brain} further incorporates explicit graph structure to model relational interactions between brain regions. Moreover, foundation models built on 3D MRI volumes have also been explored to learn transferable representations \citep{yang2025adfound}.  Despite these advances, existing methods either overlook higher-order graph structure or adopt generic graph pretraining strategies that may not fully respect the brain global organization.

\subsection{Graph Pretraining and Graph Foundation Models}

Graph pretraining \citep{hu2019strategies} is dominated by two paradigms: graph contrastive learning (GCL) \citep{you2020graph} and graph masked autoencoder (GMAE) \citep{hou2022graphmae}. GCL learns representations by maximizing agreement between augmented graph views, whereas GMAE reconstructs masked nodes or edges from partial observations. Although effective on generic graphs, both paradigms rely on naive node- or edge-level perturbations that may distort global connectivity patterns, especially in brain graphs and hypergraphs. In this work, we address these limitations by introducing a graph diffusion strategies that improves both contrastive and masking-based graph pretraining in a structure-aware manner.

\subsection{Graph Diffusion}

Graph diffusion models \citep{gasteiger2019diffusion, zhao2021adaptive, chamberlain2021grand} describe how information propagates over a graph through its topology and have been shown to improve graph representation learning. Common diffusion kernels \citep{gasteiger2019diffusion}, including the heat kernel, personalized PageRank (PPR) kernel, and random walk–based kernels, have been widely used to enhance graph convolutional networks. More recently, diffusion has also been integrated into transformer-based architectures, such as Graph Diffusion Transformer (GDT) \citep{wu2023difformer, wu2025supercharging}, where diffusion kernels are used to inject multi-hop structural bias into self-attention. However, existing studies primarily apply graph diffusion to convolutional or transformer-based models in supervised learning settings. \citep{gasteiger2019diffusion, zhao2021adaptive}, leaving its potential in self-supervised graph and hypergraph pretraining largely unexplored.

\section{Preliminaries}

\begin{figure*}[htbp]
\centering
	\includegraphics[width=17.2cm]{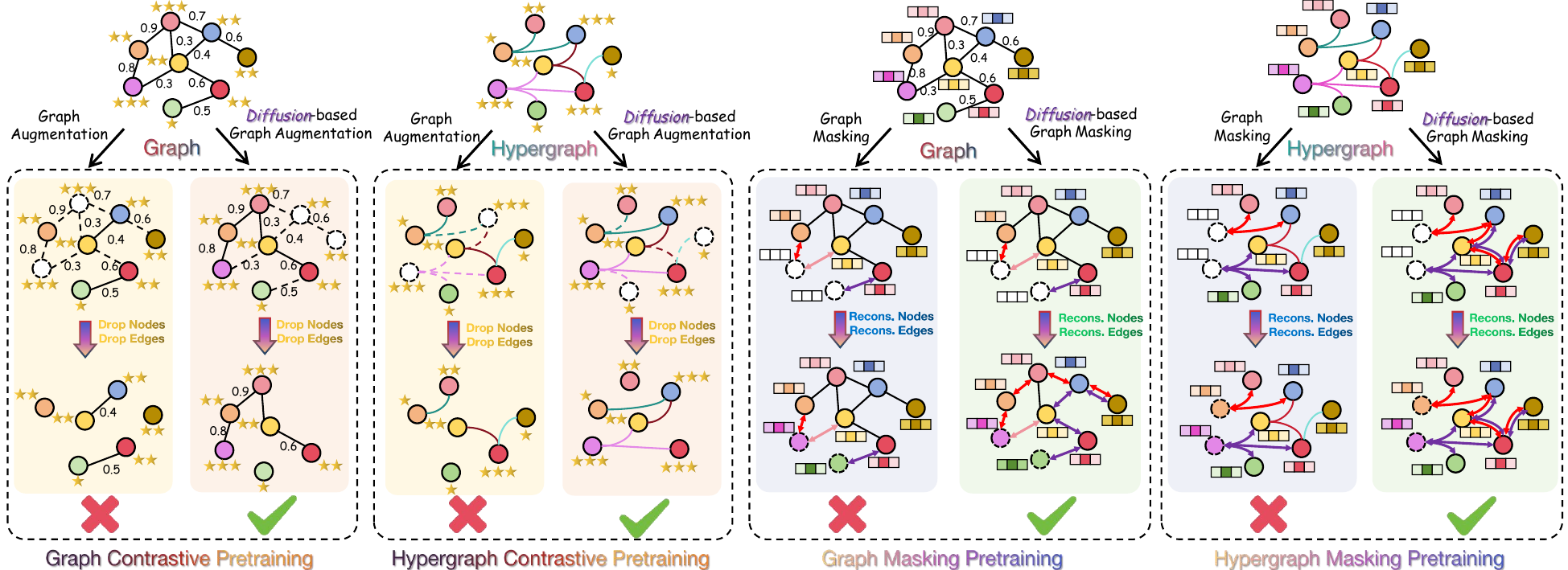}
    \vspace{-15pt}
	\caption{Diffusion-based graph and hypergraph dropping and masking strategies for augmentation. 
The figure contrasts conventional random drop/mask strategies (\ding{55}) 
with the proposed diffusion-guided approaches (\ding{51}) for both graph and 
hypergraph pretraining. Stars indicate node importance, and edge numbers denote 
weights in $[0,1]$. In GCL, nodes and edges are dropped to generate augmented 
views, while in GMAE, nodes and their features are masked and reconstructed.
Random strategies in GCL/GMAE may cause over- or under-perturbation, harming semantics or contrastiveness (blank nodes and dashed edges).
Diffusion adaptively balances perturbation strength.
Moreover, unlike local-only reconstruction in GMAE/HGMAE with edges but no arrows, diffusion enables global information aggregation across long-range nodes (bidirectional arrows).
}
	\label{model2}
    \vspace{-6pt}
\end{figure*}

\subsection{Definition of Graph and Hypergraph}

A graph \citep{kipf2016semi} is formally defined as 
\begin{equation}
G = (V, E, X)
\end{equation}
where $V = \{v_{1}, v_{2}, \dots, v_{N}\}$ denotes the set of $N$ nodes, 
$E \subseteq V \times V$ denotes the set of pairwise edges, 
and 
$X = \{x_{1}, x_{2}, \dots, x_{N}\}^\top \in \mathbb{R}^{N \times d}$
represents the node feature matrix, with $x_i \in \mathbb{R}^d$ denoting the feature vector associated with node $v_i$.
The structure of a graph can be equivalently represented by an adjacency matrix 
\begin{equation}
A \in \{0,1\}^{N\times N}, \qquad
A_{ij} = 
\begin{cases}
1, & (v_i,v_j)\in E, \\
0, & \text{otherwise}.
\end{cases}
\end{equation}
In contrast, a hypergraph \citep{feng2019hypergraph} generalizes the concept of graph edges by allowing each hyperedge to connect more than two nodes simultaneously. 
Formally, a hypergraph is defined as
\begin{equation}
\mathcal{H} = (V, \mathcal{E}, X)
\end{equation}
where $V = \{v_{1}, v_{2}, \dots, v_{N}\}$ is the set of nodes, 
$\mathcal{E} = \{e_{1}, e_{2}, \dots, e_{M}\}$ is the set of hyperedges with each $e_{m} \subseteq V$, 
and $X \in \mathbb{R}^{N \times d}$ denotes the node feature matrix shared across all hyperedges.
The incidence relation between nodes and hyperedges is encoded by an incidence matrix
\begin{equation}
I \in \{0,1\}^{N\times M}, \qquad
I_{im} = 
\begin{cases}
1, & v_i \in e_m, \\
0, & \text{otherwise}.
\end{cases}
\end{equation}
Thus, a graph can be regarded as a special case of a hypergraph where each hyperedge connects exactly two nodes.

\subsection{Graph Diffusion Mechanisms}

Graph diffusion \citep{gasteiger2019diffusion} provides a principled mechanism for propagating information over relational structures by repeatedly mixing node features according to the underlying topology.
Given an undirected graph with adjacency matrix $A$ and degree matrix $D$, we define the normalized transition matrix as $P = D^{-1}(A + I)$, where $I$ denotes the identity matrix accounting for self-loops.
We consider three representative diffusion kernels \citep{gasteiger2019diffusion} commonly used in graph learning,
\begin{equation}
\label{kernel}
\small
\begin{aligned}
K_{\mathrm{RW}}   = \sum_{k=0}^{T}& \lambda^k P^k, \quad
K_{\mathrm{PPR}}  = \alpha \sum_{k=0}^{T} (1-\alpha)^k P^k, \\
K_{\mathrm{Heat}} &= e^{-tL} \;\approx\; \sum_{k=0}^{T} \frac{(-t)^k}{k!} L^k,
\end{aligned}
\end{equation}
where $\lambda,\alpha \in (0,1)$ control the decay of long-range propagation, $t$ denotes the diffusion time scale, and 
$L = I - D^{-\tfrac{1}{2}} (A + I) D^{-\tfrac{1}{2}}$ is the normalized graph Laplacian.

Despite their different formulations, all three kernels can be interpreted as weighted combinations of multi-hop walks, enabling structure-aware information propagation beyond local neighborhoods.
Random walk diffusion aggregates fixed-length paths, PPR emphasizes locality through restart probability, and the heat kernel models continuous-time diffusion with global coverage.
Importantly, the same diffusion framework naturally extends to hypergraphs by replacing the graph adjacency matrix $A$ (or Laplacian $L$) with their hypergraph counterparts induced by the incidence matrix.
This unified formulation allows diffusion kernels to operate consistently on both graphs and hypergraphs, serving as a fundamental operator for diffusion-aware augmentation, reconstruction, and readout in our pretraining framework.

\subsection{Graph Pretraining Paradigms}

Given a brain graph or hypergraph constructed as described above, self-supervised pretraining aims to learn transferable representations without task-specific labels.
We denote the structural operator by $\mathcal{S}$, where $\mathcal{S}=A$ for graphs and $\mathcal{S}=\mathcal{H}$ for hypergraphs.
A shared encoder $f_{\theta}(\cdot)$ maps node features and structure into latent node embeddings,
\begin{equation}
Z = f_{\theta}(X, \mathcal{S}) \in \mathbb{R}^{N \times d}.
\end{equation}
\paragraph{Contrastive Learning Pretraining.}
Contrastive pretraining constructs two stochastic views of the same instance via structure-preserving transformations $\mathcal{T}(.)$,
\begin{equation}
(X^{(1)}, \mathcal{S}^{(1)}) \sim \mathcal{T}_1(X, \mathcal{S}), \quad
(X^{(2)}, \mathcal{S}^{(2)}) \sim \mathcal{T}_2(X, \mathcal{S}).
\end{equation}
Then the encoder produces view-specific embeddings $g$, which are summarized into instance-level representations by a readout function $\mathcal{R}(\cdot)$,
\begin{equation}
g^{(1)} = \mathcal{R}(f_{\theta}(X^{(1)}, \mathcal{S}^{(1)})), \quad
g^{(2)} = \mathcal{R}(f_{\theta}(X^{(2)}, \mathcal{S}^{(2)})).
\end{equation}
We adopt the standard InfoNCE \citep{you2020graph} loss function to maximize agreement between positive pairs while contrasting in-batch negatives,
\begin{equation}
\mathcal{L}_{\mathrm{GCL}}
=
-\log
\frac{\exp(\mathrm{sim}(g^{(1)}, g^{(2)})/\tau)}
{\sum\limits_{g' \in \mathcal{B}} \exp(\mathrm{sim}(g^{(1)}, g')/\tau)},
\end{equation}
where $\mathrm{sim}(\cdot,\cdot)$ denotes cosine similarity, $\tau$ is a temperature parameter, and $\mathcal{B}$ denotes the set of instance representations in the batch.

\paragraph{Masked Autoencoder Pretraining.}
Masked autoencoder randomly masks a subset of nodes $\mathcal{M}\subseteq V$ with ratio $\rho$ to form corrupted inputs $\tilde{X}$.
The encoder $f_{\theta}(\cdot)$ and a lightweight decoder $g_{\phi}(\cdot)$ are used to reconstruct masked node features,
\begin{equation}
Z = f_{\theta}(\tilde{X}, \mathcal{S}), \qquad
\hat{X}_{\mathcal{M}} = g_{\phi}(Z_{\mathcal{M}}),
\end{equation}
and the reconstruction loss is computed over masked nodes,
\begin{equation}
\mathcal{L}_{\mathrm{GMAE}}
=
\frac{1}{|\mathcal{M}|}
\sum_{i \in \mathcal{M}}
\lVert X_i - \hat{X}_i \rVert_2^2.
\end{equation}

\section{Methodology}

\subsection{Diffusion-Guided Contrastive Pretraining}
\label{sec:diff_gcl}

Conventional graph and hypergraph contrastive pretraining paradigms often build contrastive views by randomly dropping nodes/edges/hyperedges.
However, brain networks encode semantics in global connectivity patterns, so random perturbations can be either overly destructive or too weak.
We propose diffusion-guided augmentations that (i) globally aggregate topology-aware context via diffusion and (ii) stochastically drop less diffusion-supported components to produce diffusion-consistent views (Alg.~\ref{alg:gcl_hgcl} in Appendix).

\paragraph{Unified Diffusion over Graphs and Hypergraphs.}
We denote the structure operator by $\mathcal{S}$ (graph: $\mathcal{S}=A$, hypergraph: $\mathcal{S}=I$) and construct a Markov transition $P(\mathcal{S})$.
For graphs, we use the row-stochastic transition
\begin{equation}
P = D^{-1}(A+I),
\label{eq:trans_graph}
\end{equation}
where $D$ is the degree matrix of $(A+I)$.
For hypergraphs, we induce a node-to-node transition via the incidence matrix $I\in\{0,1\}^{N\times M}$:
\begin{equation}
P = D_v^{-1} I W D_e^{-1} I^\top,
\label{eq:trans_hyper}
\end{equation}
where $W\in\mathbb{R}^{M\times M}$ is a diagonal hyperedge weight matrix,
$(D_v)_{ii}=\sum_{m} I_{im} w_m$, and $(D_e)_{mm}=\sum_i I_{im}$.
Given $P$, we define a multi-hop diffusion kernel $K(.)$ in Eq. (\ref{kernel})
and obtain contextualized node features $X^{\mathrm{diff}}$ by diffusion:
\begin{equation}
X^{\mathrm{diff}} = KX.
\label{eq:diffuse_feature}
\end{equation}

\paragraph{Diffusion-Guided Dropping Augmentation.}
Firstly, we score node importance using the diffusion energy \citep{wu2023difformer} of diffused features:
\begin{equation}
s_i^{(v)} = \big\| (X^{\mathrm{diff}})_i \big\|_2.
\label{eq:node_score}
\end{equation}
Nodes with smaller diffusion energy $s_i^{(v)}$ are less supported by global diffusion and are safer to drop.
We then map scores to \emph{moderate} diffusion-aware drop probabilities via
\begin{equation}
p_i^{\mathrm{drop}} = \Psi\!\left(s_i^{(v)}\right),
\quad
\mathcal{M}^{(1)},\mathcal{M}^{(2)} \sim \mathrm{Bernoulli}(1-p^{\mathrm{drop}}),
\label{eq:drop_prob_mask}
\end{equation}
where $\Psi(\cdot)$ is a monotonic function that increases the drop probability for less important components (Alg.~\ref{alg:gcl_hgcl} in Appendix). $\mathcal{M}^{(1)}$ and $\mathcal{M}^{(2)}$ denote two independently sampled drop masks for constructing two contrastive views. 

To make structural perturbations topology-aware, we first compute a diffusion-enhanced connectivity matrix
\begin{equation}
\widetilde{A} = K A K^\top,
\end{equation}
which captures global, multi-hop interactions between nodes.
Based on $\widetilde{A}$, we define diffusion-guided importance scores for both graph edges and hyperedges as
\begin{equation}
\begin{aligned}
s_{ij}^{\mathrm{graph}} &= \widetilde{A}{ij}, \quad
s_m^{\mathrm{hyperG}} = u_m^\top \widetilde{A} , u_m, \\
\mathcal{S}^{(v)} &= \mathcal{S} \odot \Psi(s), \quad v\in{1,2}.
\end{aligned}
\end{equation}
where $s{ij}^{\mathrm{graph}}$ measures the importance of edge $(i,j)$ in a graph, and $s_m^{\mathrm{hyperG}}$ quantifies the structural coherence of hyperedge $e_m$ by aggregating diffusion-supported connectivity among its incident nodes with indicator vector $u_m$.
Here, $\Psi(\cdot)$ denotes a monotonic mapping that converts importance scores into stochastic structure masks, thereby controlling which edges or hyperedges in $\mathcal{S}$ are retained or removed.

Finally, we generate two augmented views by applying diffusion-guided node and structure dropping on the diffused features:
\begin{equation}
(X^{(v)}, \mathcal{S}^{(v)}) = \mathrm{Drop}(X^{\mathrm{diff}}, \mathcal{S}, \mathcal{M}^{(v)}),
\quad v\in{1,2}.
\label{eq:two_views}
\end{equation}
This procedure yields stochastic yet semantics-preserving views that avoid destructive random perturbations while maintaining sufficient diversity for contrastive supervision on brain graphs and hypergraphs.

\paragraph{Diffusion-Based Graph Readout.}
Conventional graph readout \citep{buterez2022graph}, including mean pooling, max pooling, and attention-based pooling (details can be founda in Appendix \ref{app:readout}), aggregate node features in a structure-agnostic manner, treating nodes as an unordered set.



To obtain stable and semantically consistent contrastive view embeddings, we propose a \emph{diffusion-based readout}.
Instead of directly pooling raw node features, we first diffuse node embeddings over the graph (or hypergraph) structure using a diffusion kernel $K$,
which encodes multi-hop information propagation (e.g., random walk (RW), personalized PageRank (PPR), or heat kernel \citep{gasteiger2019diffusion}.
Given node embeddings $X\in\mathbb{R}^{N\times d}$, diffusion is performed as
\begin{equation}
\tilde{X} = KX,
\qquad
\tilde{x}_i = \sum_{j=1}^{N} K_{ij} x_j,
\end{equation}
so that each node integrates global structural context before aggregation.
A permutation-invariant pooling is then applied to obtain the graph-level representation used in contrastive learning:
\begin{equation}
\mathcal{R}_{\mathrm{diff}} = \mathrm{Pool}(\tilde{X})
= \frac{1}{N}\mathbf{I}^\top KX,
\end{equation}
where $\mathbf{I}\in\mathbb{R}^{N}$ denotes the all-ones vector.
By performing topology-aware diffusion prior to pooling, the proposed readout yields more robust and globally consistent embeddings for contrastive views, facilitating effective contrastive pretraining on brain graphs and hypergraphs.

\paragraph{Graph Contrastive Loss.}
Given two diffusion-consistent augmented views
$\{(X^{(v)}, \mathcal{S}^{(v)})\}_{v=1}^{2}$,
we obtain graph-level embeddings via a shared encoder $f_\theta(.)$ followed by diffusion-based readout:
\begin{equation}
g^{(v)} = \mathcal{R}_{\mathrm{diff}}\!\left(K f_\theta(X^{(v)}, \mathcal{S}^{(v)})\right).
\label{eq:diff_readout_gcl}
\end{equation}
Finally, contrastive learning is performed by maximizing agreement between the two
diffusion-consistent views using the NT-Xent objective:
\begin{equation}
\mathcal{L}_{\mathrm{GCL}}
=
-\log
\frac{
\exp\!\left(\mathrm{sim}(g^{(1)}, g^{(2)}) / \tau \right)
}{
\sum_{g' \neq g^{(2)}}
\exp\!\left(\mathrm{sim}(g^{(1)}, g') / \tau \right)
}.
\label{eq:diff_gcl_loss}
\end{equation}

\subsection{Diffusion-Guided Masked Autoencoder Pretraining}
\label{sec:diff_gmae}

Previous Graph Masked Autoencoders reconstruct masked nodes or edges mainly from local neighborhoods \citep{hou2022graphmae}, which can be fragile for brain graphs and hypergraphs with long-range interactions \citep{yu2024long} and higher-order dependencies \citep{han2025hypergraph}.
We therefore enhance masked autoencoder with diffusion to supply global structural context for reconstruction.

\paragraph{Diffusion-Guided Masking.}
Following the diffusion-guided dropping strategy in contrastive pretraining, we guide node masking using diffusion-informed importance scores (Alg.~\ref{alg:gmae_hgmae}).
Nodes that are weakly supported under global diffusion are more likely to be masked, while globally important nodes are preserved with higher probability.
Since masking is conceptually analogous to dropping but without removing nodes from the graph, we omit repeated formulations for brevity.

\paragraph{Diffusion-Propagated Feature Reconstruction.}
Let $\mathcal{M}\subseteq V$ denote the set of masked nodes with masking ratio $\rho$, and let $\tilde{X}$ be the masked input features.
We first diffuse the masked features to inject global context:
\begin{equation}
\tilde{X}^{\mathrm{diff}} = K \tilde{X}.
\end{equation}
The diffusion-enhanced features are encoded as node-level latent representations:
\begin{equation}
Z = f_\theta(\tilde{X}^{\mathrm{diff}}, \mathcal{S}),
\end{equation}
where $Z=\{z_i\}_{i\in V}$ and $Z_{\mathcal{M}}=\{z_i\}_{i\in\mathcal{M}}$ denotes the masked-node subset.
We further diffuse the latent representations:
\begin{equation}
\hat{Z} = K Z,
\qquad
\hat{X}_{\mathcal{M}} = g_\phi(\hat{Z}_{\mathcal{M}}),
\end{equation}
where $\hat{Z}$ denotes the diffusion-propagated latent embeddings.
The reconstruction loss is
\begin{equation}
\mathcal{L}_{\mathrm{node}}
=
\frac{1}{|\mathcal{M}|}
\sum_{i\in\mathcal{M}}
\| X_i - \hat{X}_i \|_2^2 .
\end{equation}

\paragraph{Diffusion-Guided Global Structure Reconstruction.}
Beyond node features, we reconstruct masked graph and hypergraph structures using
diffusion-consistent targets.

Let $z_i$ denote the embedding of node $v_i$.
We predict a soft diffusion-consistent connectivity score between nodes via
\begin{equation}
\widehat{\widetilde{A}}_{ij} = \sigma(z_i^\top z_j),
\end{equation}
where $\sigma(\cdot)$ is the sigmoid function.
Instead of the original adjacency, we use a diffusion-enhanced adjacency $\widetilde{A}\in[0,1]^{N\times N}$
as the reconstruction target.
The edge loss is computed only on masked edges $\mathcal{E}_{\mathrm{mask}}$:
\begin{equation}
\mathcal{L}_{\mathrm{edge}}
=
\frac{1}{|\mathcal{E}_{\mathrm{mask}}|}
\sum_{(i,j)\in\mathcal{E}_{\mathrm{mask}}}
\mathrm{BCE}\!\left(\widehat{\widetilde{A}}_{ij}, \widetilde{A}_{ij}\right),
\end{equation}
where $\mathrm{BCE}(\cdot,\cdot)$ denotes binary cross-entropy.
Diffusion targets provide smoother supervision with multi-hop evidence and are less sensitive to local noise.

For a hyperedge $e_m$ with indicator vector $u_m$, we aggregate incident node embeddings by
\begin{equation}
\widehat{s}_m
=
\sigma\!\left(\mathrm{Pool}(\{z_i \mid u_{m,i}=1\})\right),
\end{equation}
and minimize
\begin{equation}
\mathcal{L}_{\mathrm{hyper}} = \sum_m \|\widehat{s}_m - s_m^{(h)}\|_2^2,
\end{equation}
where $s_m^{(h)}$ is the diffusion-induced hyperedge target.

The final diffusion-enhanced GMAE loss is
\begin{equation}
\mathcal{L}_{\mathrm{GMAE}}^{\mathrm{diff}}
=
\mathcal{L}_{\mathrm{node}}
+
\eta\,\mathcal{L}_{\mathrm{struct}},
\end{equation}
$\mathcal{L}_{\mathrm{struct}}$ denotes $\mathcal{L}_{\mathrm{edge}}$ for graphs or $\mathcal{L}_{\mathrm{hyper}}$ for hypergraphs.

\section{Experiments}

\subsection{Comparative Experiment Analysis}

\begin{table*}[ht]
\scriptsize
\renewcommand\arraystretch{.9}
\setlength{\tabcolsep}{.1pt}
\centering
\caption{Comparison among different methods on diagnoses of multiple brain disorders based on the Schaefer atlas of 100 ROIs. Methods are grouped into traditional deep learning models, traditional diffusion models, pretrained models, and diffusion-based pretrained models.}
\label{comparison}
\vspace{-4pt}
\begin{tabular}{c|c|ccccccccccccccccccc}
\toprule
\multirow{2}{*}{Method} & \multirow{2}{*}{Pretrained} 

& \multicolumn{3}{c}{ADHD200 (ADHD)} & \multicolumn{3}{c}{OASIS3 (DM)} & \multicolumn{3}{c}{ADNI 2 (AD)}  & \multicolumn{3}{c}{HBN (MDD)}  & \multicolumn{3}{c}{SubMex\_CUD (CUD)}  & \multicolumn{3}{c}{UCLA\_CNP (BP)}   \\ \cmidrule(lr){3-5} \cmidrule(lr){6-8} \cmidrule(lr){9-11} \cmidrule(lr){12-14} \cmidrule(lr){15-17}   \cmidrule(lr){18-20} 
& & AUC & ACC   & F1  & AUC & ACC   & F1   & AUC & ACC   & F1 & AUC & ACC   & F1 & AUC & ACC   & F1 & AUC & ACC   & F1  \\ \midrule

\rowcolor{gray!20}BrainGNN &\XSolidBrush  
&61.3\textsubscript{\fontsize{4pt}{4pt}\selectfont ±3.1}  
&64.4\textsubscript{\fontsize{4pt}{4pt}\selectfont ±2.6} 
&62.6\textsubscript{\fontsize{4pt}{4pt}\selectfont ±2.8} 

&65.1\textsubscript{\fontsize{4pt}{4pt}\selectfont ±2.7}  
&67.2\textsubscript{\fontsize{4pt}{4pt}\selectfont ±3.2} 
&68.3\textsubscript{\fontsize{4pt}{4pt}\selectfont ±2.7} 

&71.1\textsubscript{\fontsize{4pt}{4pt}\selectfont ±1.6}  
&71.2\textsubscript{\fontsize{4pt}{4pt}\selectfont ±2.9}  
&74.5\textsubscript{\fontsize{4pt}{4pt}\selectfont ±3.4}

&69.0\textsubscript{\fontsize{4pt}{4pt}\selectfont ±3.1}  
&74.1\textsubscript{\fontsize{4pt}{4pt}\selectfont ±2.2} 
&75.3\textsubscript{\fontsize{4pt}{4pt}\selectfont ±2.7} 

&63.9\textsubscript{\fontsize{4pt}{4pt}\selectfont ±3.4}  
&66.1\textsubscript{\fontsize{4pt}{4pt}\selectfont ±3.2} 
&64.3\textsubscript{\fontsize{4pt}{4pt}\selectfont ±2.2}

&66.0\textsubscript{\fontsize{4pt}{4pt}\selectfont ±3.4}  
&71.3\textsubscript{\fontsize{4pt}{4pt}\selectfont ±2.8} 
&73.5\textsubscript{\fontsize{4pt}{4pt}\selectfont ±3.0} 

\\

\rowcolor{gray!20}BrainNetTF & \XSolidBrush  
&64.7\textsubscript{\fontsize{4pt}{4pt}\selectfont ±2.3}  
&66.0\textsubscript{\fontsize{4pt}{4pt}\selectfont ±3.2}  
&65.8\textsubscript{\fontsize{4pt}{4pt}\selectfont ±2.2} 

&68.3\textsubscript{\fontsize{4pt}{4pt}\selectfont ±2.5}  
&70.7\textsubscript{\fontsize{4pt}{4pt}\selectfont ±2.7} 
&71.5\textsubscript{\fontsize{4pt}{4pt}\selectfont ±2.2} 

&75.9\textsubscript{\fontsize{4pt}{4pt}\selectfont ±2.6}  
&77.6\textsubscript{\fontsize{4pt}{4pt}\selectfont ±2.0}  
&80.3\textsubscript{\fontsize{4pt}{4pt}\selectfont ±3.1} 

&74.2\textsubscript{\fontsize{4pt}{4pt}\selectfont ±2.4}  
&76.3\textsubscript{\fontsize{4pt}{4pt}\selectfont ±2.1} 
&78.5\textsubscript{\fontsize{4pt}{4pt}\selectfont ±3.1} 

&68.3\textsubscript{\fontsize{4pt}{4pt}\selectfont ±2.9}  
&69.1\textsubscript{\fontsize{4pt}{4pt}\selectfont ±3.4}  
&66.7\textsubscript{\fontsize{4pt}{4pt}\selectfont ±2.6} 

&68.1\textsubscript{\fontsize{4pt}{4pt}\selectfont ±3.1}  
&73.2\textsubscript{\fontsize{4pt}{4pt}\selectfont ±3.4}  
&75.4\textsubscript{\fontsize{4pt}{4pt}\selectfont ±2.5} 
 
\\

 



\rowcolor{gray!10}GDC&\XSolidBrush   

&62.1\textsubscript{\fontsize{4pt}{4pt}\selectfont ±2.5}  
&64.4\textsubscript{\fontsize{4pt}{4pt}\selectfont ±2.3} 
&63.6\textsubscript{\fontsize{4pt}{4pt}\selectfont ±2.9} 

&66.7\textsubscript{\fontsize{4pt}{4pt}\selectfont ±3.1}  
&67.9\textsubscript{\fontsize{4pt}{4pt}\selectfont ±2.7} 
&68.7\textsubscript{\fontsize{4pt}{4pt}\selectfont ±2.4} 

&72.5\textsubscript{\fontsize{4pt}{4pt}\selectfont ±2.2}  
&74.7\textsubscript{\fontsize{4pt}{4pt}\selectfont ±2.6}  
&76.8\textsubscript{\fontsize{4pt}{4pt}\selectfont ±3.5} 

&72.6\textsubscript{\fontsize{4pt}{4pt}\selectfont ±3.0}  
&73.9\textsubscript{\fontsize{4pt}{4pt}\selectfont ±2.8} 
&75.5\textsubscript{\fontsize{4pt}{4pt}\selectfont ±2.8}

&65.8\textsubscript{\fontsize{4pt}{4pt}\selectfont ±3.2}  
&66.3\textsubscript{\fontsize{4pt}{4pt}\selectfont ±2.5} 
&65.3\textsubscript{\fontsize{4pt}{4pt}\selectfont ±2.7} 

&67.3\textsubscript{\fontsize{4pt}{4pt}\selectfont ±3.0}  
&70.8\textsubscript{\fontsize{4pt}{4pt}\selectfont ±2.8} 
&71.6\textsubscript{\fontsize{4pt}{4pt}\selectfont ±2.3}

\\ 

\rowcolor{gray!10}Adaptive GDC &\XSolidBrush   

&63.7\textsubscript{\fontsize{4pt}{4pt}\selectfont ±2.7}  
&66.6\textsubscript{\fontsize{4pt}{4pt}\selectfont ±1.6} 
&65.6\textsubscript{\fontsize{4pt}{4pt}\selectfont ±2.4} 

&67.3\textsubscript{\fontsize{4pt}{4pt}\selectfont ±2.9}  
&69.6\textsubscript{\fontsize{4pt}{4pt}\selectfont ±3.3} 
&70.4\textsubscript{\fontsize{4pt}{4pt}\selectfont ±3.2} 

&73.4\textsubscript{\fontsize{4pt}{4pt}\selectfont ±3.0}  
&75.8\textsubscript{\fontsize{4pt}{4pt}\selectfont ±2.8}  
&78.6\textsubscript{\fontsize{4pt}{4pt}\selectfont ±3.3} 

&74.2\textsubscript{\fontsize{4pt}{4pt}\selectfont ±3.0}  
&75.7\textsubscript{\fontsize{4pt}{4pt}\selectfont ±2.8} 
&77.5\textsubscript{\fontsize{4pt}{4pt}\selectfont ±3.2}

&66.9\textsubscript{\fontsize{4pt}{4pt}\selectfont ±2.7}  
&68.7\textsubscript{\fontsize{4pt}{4pt}\selectfont ±2.4} 
&66.8\textsubscript{\fontsize{4pt}{4pt}\selectfont ±1.9} 

&68.5\textsubscript{\fontsize{4pt}{4pt}\selectfont ±2.6}  
&72.3\textsubscript{\fontsize{4pt}{4pt}\selectfont ±2.7} 
&73.6\textsubscript{\fontsize{4pt}{4pt}\selectfont ±2.9} 

\\ 

\rowcolor{gray!10}GDT &\XSolidBrush   

&65.0\textsubscript{\fontsize{4pt}{4pt}\selectfont ±2.0}  
&66.8\textsubscript{\fontsize{4pt}{4pt}\selectfont ±2.2} 
&66.9\textsubscript{\fontsize{4pt}{4pt}\selectfont ±2.3} 

&68.1\textsubscript{\fontsize{4pt}{4pt}\selectfont ±2.3}  
&70.2\textsubscript{\fontsize{4pt}{4pt}\selectfont ±2.5} 
&70.7\textsubscript{\fontsize{4pt}{4pt}\selectfont ±2.8} 

&76.5\textsubscript{\fontsize{4pt}{4pt}\selectfont ±1.8}  
&77.7\textsubscript{\fontsize{4pt}{4pt}\selectfont ±2.6} 
&79.5\textsubscript{\fontsize{4pt}{4pt}\selectfont ±3.2}

&75.6\textsubscript{\fontsize{4pt}{4pt}\selectfont ±1.9}  
&77.2\textsubscript{\fontsize{4pt}{4pt}\selectfont ±2.3} 
&79.3\textsubscript{\fontsize{4pt}{4pt}\selectfont ±2.6} 

&67.6\textsubscript{\fontsize{4pt}{4pt}\selectfont ±2.3}  
&70.4\textsubscript{\fontsize{4pt}{4pt}\selectfont ±2.1} 
&67.5\textsubscript{\fontsize{4pt}{4pt}\selectfont ±3.1} 

&69.4\textsubscript{\fontsize{4pt}{4pt}\selectfont ±1.7}  
&73.2\textsubscript{\fontsize{4pt}{4pt}\selectfont ±2.3} 
&75.3\textsubscript{\fontsize{4pt}{4pt}\selectfont ±3.1} 

\\

\rowcolor{gray!20}BrainLM &\Checkmark  
&67.6\textsubscript{\fontsize{4pt}{4pt}\selectfont ±1.9}  
&69.1\textsubscript{\fontsize{4pt}{4pt}\selectfont ±1.7}  
&68.7\textsubscript{\fontsize{4pt}{4pt}\selectfont ±2.8} 

&71.4\textsubscript{\fontsize{4pt}{4pt}\selectfont ±2.2}  
&73.6\textsubscript{\fontsize{4pt}{4pt}\selectfont ±2.1} 
&73.7\textsubscript{\fontsize{4pt}{4pt}\selectfont ±2.5} 

&78.3\textsubscript{\fontsize{4pt}{4pt}\selectfont ±1.7}  
&79.7\textsubscript{\fontsize{4pt}{4pt}\selectfont ±2.5}  
&82.6\textsubscript{\fontsize{4pt}{4pt}\selectfont ±2.8} 

&77.8\textsubscript{\fontsize{4pt}{4pt}\selectfont ±2.0}  
&80.7\textsubscript{\fontsize{4pt}{4pt}\selectfont ±2.1}  
&81.3\textsubscript{\fontsize{4pt}{4pt}\selectfont ±2.8}

&68.2\textsubscript{\fontsize{4pt}{4pt}\selectfont ±2.3}  
&72.5\textsubscript{\fontsize{4pt}{4pt}\selectfont ±2.1}  
&70.4\textsubscript{\fontsize{4pt}{4pt}\selectfont ±2.6} 

&71.5\textsubscript{\fontsize{4pt}{4pt}\selectfont ±2.2}  
&74.9\textsubscript{\fontsize{4pt}{4pt}\selectfont ±1.8}  
&76.4\textsubscript{\fontsize{4pt}{4pt}\selectfont ±1.5} 

\\

\rowcolor{gray!20}BrainMass & \Checkmark  
&67.0\textsubscript{\fontsize{4pt}{4pt}\selectfont ±2.3}  
&67.5\textsubscript{\fontsize{4pt}{4pt}\selectfont ±2.0}  
&68.6\textsubscript{\fontsize{4pt}{4pt}\selectfont ±2.3} 

&70.3\textsubscript{\fontsize{4pt}{4pt}\selectfont ±2.2}  
&72.5\textsubscript{\fontsize{4pt}{4pt}\selectfont ±2.5} 
&73.6\textsubscript{\fontsize{4pt}{4pt}\selectfont ±2.9} 

&77.8\textsubscript{\fontsize{4pt}{4pt}\selectfont ±2.3}  
&80.6\textsubscript{\fontsize{4pt}{4pt}\selectfont ±2.7}  
&82.7\textsubscript{\fontsize{4pt}{4pt}\selectfont ±2.7} 

&76.9\textsubscript{\fontsize{4pt}{4pt}\selectfont ±2.4}  
&81.5\textsubscript{\fontsize{4pt}{4pt}\selectfont ±2.8}  
&81.9\textsubscript{\fontsize{4pt}{4pt}\selectfont ±2.5}  

&68.0\textsubscript{\fontsize{4pt}{4pt}\selectfont ±2.0}  
&70.9\textsubscript{\fontsize{4pt}{4pt}\selectfont ±2.3}  
&69.6\textsubscript{\fontsize{4pt}{4pt}\selectfont ±2.0}

&70.8\textsubscript{\fontsize{4pt}{4pt}\selectfont ±2.2}  
&73.6\textsubscript{\fontsize{4pt}{4pt}\selectfont ±2.3}  
&76.0\textsubscript{\fontsize{4pt}{4pt}\selectfont ±1.9}

\\

\rowcolor{gray!20}Brain-JEPA & \Checkmark  
&69.8\textsubscript{\fontsize{4pt}{4pt}\selectfont ±1.9}  &71.6\textsubscript{\fontsize{4pt}{4pt}\selectfont ±2.0} 
&70.2\textsubscript{\fontsize{4pt}{4pt}\selectfont ±2.1} 

&72.4\textsubscript{\fontsize{4pt}{4pt}\selectfont ±2.2}  
&74.5\textsubscript{\fontsize{4pt}{4pt}\selectfont ±2.6} 
&75.5\textsubscript{\fontsize{4pt}{4pt}\selectfont ±2.4} 

&79.1\textsubscript{\fontsize{4pt}{4pt}\selectfont ±2.2}  &81.6\textsubscript{\fontsize{4pt}{4pt}\selectfont ±1.8}  
&84.2\textsubscript{\fontsize{4pt}{4pt}\selectfont ±2.4} 

&79.3\textsubscript{\fontsize{4pt}{4pt}\selectfont ±1.7}  &82.8\textsubscript{\fontsize{4pt}{4pt}\selectfont ±1.3}  
&82.4\textsubscript{\fontsize{4pt}{4pt}\selectfont ±2.8}  

&70.1\textsubscript{\fontsize{4pt}{4pt}\selectfont ±2.0}  
&74.9\textsubscript{\fontsize{4pt}{4pt}\selectfont ±2.3}  
&71.5\textsubscript{\fontsize{4pt}{4pt}\selectfont ±2.2} 

&73.7\textsubscript{\fontsize{4pt}{4pt}\selectfont ±2.2}  
&75.2\textsubscript{\fontsize{4pt}{4pt}\selectfont ±2.1}  
&77.3\textsubscript{\fontsize{4pt}{4pt}\selectfont ±2.7}

\\

\midrule

\rowcolor{gray!10} BrainGFM   & \Checkmark
&70.3\textsubscript{\fontsize{4pt}{4pt}\selectfont ±1.6}  &70.5\textsubscript{\fontsize{4pt}{4pt}\selectfont ±1.5}  
&70.1\textsubscript{\fontsize{4pt}{4pt}\selectfont ±1.9}  

&72.9\textsubscript{\fontsize{4pt}{4pt}\selectfont ±2.7}  
&75.4\textsubscript{\fontsize{4pt}{4pt}\selectfont ±3.2} 
&76.2\textsubscript{\fontsize{4pt}{4pt}\selectfont ±2.7} 

&80.3\textsubscript{\fontsize{4pt}{4pt}\selectfont ±2.6}  &81.9\textsubscript{\fontsize{4pt}{4pt}\selectfont ±2.2}  
&84.7\textsubscript{\fontsize{4pt}{4pt}\selectfont ±2.4} 

&79.9\textsubscript{\fontsize{4pt}{4pt}\selectfont ±1.6}  &82.0\textsubscript{\fontsize{4pt}{4pt}\selectfont ±1.7}  
&84.8\textsubscript{\fontsize{4pt}{4pt}\selectfont ±2.9}  

&71.1\textsubscript{\fontsize{4pt}{4pt}\selectfont ±1.6}  
&74.6\textsubscript{\fontsize{4pt}{4pt}\selectfont ±1.8}  
&72.2\textsubscript{\fontsize{4pt}{4pt}\selectfont ±1.9} 

&73.5\textsubscript{\fontsize{4pt}{4pt}\selectfont ±1.8}  
&76.3\textsubscript{\fontsize{4pt}{4pt}\selectfont ±1.9} 
&78.6\textsubscript{\fontsize{4pt}{4pt}\selectfont ±2.4} 

\\

\rowcolor{gray!10}Pretrained GDT & \Checkmark  

&71.4\textsubscript{\fontsize{4pt}{4pt}\selectfont ±1.7}  
&72.1\textsubscript{\fontsize{4pt}{4pt}\selectfont ±2.0}  
&71.7\textsubscript{\fontsize{4pt}{4pt}\selectfont ±2.1}  

&73.7\textsubscript{\fontsize{4pt}{4pt}\selectfont ±2.3}  
&76.8\textsubscript{\fontsize{4pt}{4pt}\selectfont ±2.4} 
&76.5\textsubscript{\fontsize{4pt}{4pt}\selectfont ±2.2} 

&82.1\textsubscript{\fontsize{4pt}{4pt}\selectfont ±1.9}  
&83.9\textsubscript{\fontsize{4pt}{4pt}\selectfont ±1.7}  
&85.6\textsubscript{\fontsize{4pt}{4pt}\selectfont ±2.2} 

&81.3\textsubscript{\fontsize{4pt}{4pt}\selectfont ±2.1}  
&83.6\textsubscript{\fontsize{4pt}{4pt}\selectfont ±2.0} 
&85.8\textsubscript{\fontsize{4pt}{4pt}\selectfont ±2.3}  

&72.6\textsubscript{\fontsize{4pt}{4pt}\selectfont ±1.6}  
&75.9\textsubscript{\fontsize{4pt}{4pt}\selectfont ±1.8} 
&74.4\textsubscript{\fontsize{4pt}{4pt}\selectfont ±2.1} 

&74.1\textsubscript{\fontsize{4pt}{4pt}\selectfont ±2.0}  
&77.9\textsubscript{\fontsize{4pt}{4pt}\selectfont ±1.4} 
&79.3\textsubscript{\fontsize{4pt}{4pt}\selectfont ±1.7} 

\\  

\rowcolor{gray!10} Brain-HyperGFM   & \Checkmark

&71.8\textsubscript{\fontsize{4pt}{4pt}\selectfont ±1.5}  
&71.9\textsubscript{\fontsize{4pt}{4pt}\selectfont ±2.0} 
&72.3\textsubscript{\fontsize{4pt}{4pt}\selectfont ±1.6}  

&74.3\textsubscript{\fontsize{4pt}{4pt}\selectfont ±1.9}  
&76.9\textsubscript{\fontsize{4pt}{4pt}\selectfont ±1.7} 
&77.3\textsubscript{\fontsize{4pt}{4pt}\selectfont ±2.3}

&82.9\textsubscript{\fontsize{4pt}{4pt}\selectfont ±1.5}  
&82.4\textsubscript{\fontsize{4pt}{4pt}\selectfont ±1.3} 
&85.2\textsubscript{\fontsize{4pt}{4pt}\selectfont ±1.7}

&81.4\textsubscript{\fontsize{4pt}{4pt}\selectfont ±1.8}  
&83.7\textsubscript{\fontsize{4pt}{4pt}\selectfont ±1.4} 
&85.2\textsubscript{\fontsize{4pt}{4pt}\selectfont ±1.8}

&72.8\textsubscript{\fontsize{4pt}{4pt}\selectfont ±1.4}  
&75.6\textsubscript{\fontsize{4pt}{4pt}\selectfont ±1.8} 
&73.2\textsubscript{\fontsize{4pt}{4pt}\selectfont ±2.2}  

&74.3\textsubscript{\fontsize{4pt}{4pt}\selectfont ±1.8}  
&78.4\textsubscript{\fontsize{4pt}{4pt}\selectfont ±1.4} 
&79.4\textsubscript{\fontsize{4pt}{4pt}\selectfont ±1.8}  

\\ 
\midrule

\rowcolor{gray!10} BrainGFM-Diff   & \Checkmark

&72.6\textsubscript{\fontsize{4pt}{4pt}\selectfont ±1.8}  
&73.1\textsubscript{\fontsize{4pt}{4pt}\selectfont ±1.4} 
&73.6\textsubscript{\fontsize{4pt}{4pt}\selectfont ±1.8}  

&75.4\textsubscript{\fontsize{4pt}{4pt}\selectfont ±2.3}  
&77.6\textsubscript{\fontsize{4pt}{4pt}\selectfont ±1.6} 
&78.7\textsubscript{\fontsize{4pt}{4pt}\selectfont ±2.2} 

&83.0\textsubscript{\fontsize{4pt}{4pt}\selectfont ±1.5}  
&82.6\textsubscript{\fontsize{4pt}{4pt}\selectfont ±1.3}  
&85.4\textsubscript{\fontsize{4pt}{4pt}\selectfont ±2.4} 

&83.1\textsubscript{\fontsize{4pt}{4pt}\selectfont ±1.6}  
&84.5\textsubscript{\fontsize{4pt}{4pt}\selectfont ±1.9}  
&86.3\textsubscript{\fontsize{4pt}{4pt}\selectfont ±2.1}

&73.7\textsubscript{\fontsize{4pt}{4pt}\selectfont ±1.6}  
&76.8\textsubscript{\fontsize{4pt}{4pt}\selectfont ±1.9}  
&75.4\textsubscript{\fontsize{4pt}{4pt}\selectfont ±3.2} 

&75.2\textsubscript{\fontsize{4pt}{4pt}\selectfont ±1.3}  
&78.8\textsubscript{\fontsize{4pt}{4pt}\selectfont ±1.5} 
&80.5\textsubscript{\fontsize{4pt}{4pt}\selectfont ±2.4}

\\

\rowcolor{gray!10} Brain-HyperGFM-Diff   & \Checkmark 

&72.8\textsubscript{\fontsize{4pt}{4pt}\selectfont ±1.4}  
&73.4\textsubscript{\fontsize{4pt}{4pt}\selectfont ±1.8} 
&\cellcolor{pink!50}73.7\textsubscript{\fontsize{4pt}{4pt}\selectfont ±2.3}

&74.8\textsubscript{\fontsize{4pt}{4pt}\selectfont ±1.9}  
&77.8\textsubscript{\fontsize{4pt}{4pt}\selectfont ±1.7} 
&\cellcolor{pink!50}78.9\textsubscript{\fontsize{4pt}{4pt}\selectfont ±2.1}

&\cellcolor{pink!50}84.0\textsubscript{\fontsize{4pt}{4pt}\selectfont ±1.5}  
&\cellcolor{pink!50}83.5\textsubscript{\fontsize{4pt}{4pt}\selectfont ±1.7} 
&\cellcolor{pink!50}86.6\textsubscript{\fontsize{4pt}{4pt}\selectfont ±2.5}

&82.9\textsubscript{\fontsize{4pt}{4pt}\selectfont ±1.6}  
&\cellcolor{pink!50}85.2\textsubscript{\fontsize{4pt}{4pt}\selectfont ±1.3} 
&86.7\textsubscript{\fontsize{4pt}{4pt}\selectfont ±2.0}

&74.0\textsubscript{\fontsize{4pt}{4pt}\selectfont ±1.5}  
&\cellcolor{pink!50}77.1\textsubscript{\fontsize{4pt}{4pt}\selectfont ±1.9} 
&75.2\textsubscript{\fontsize{4pt}{4pt}\selectfont ±1.5}

&\cellcolor{pink!50}76.6\textsubscript{\fontsize{4pt}{4pt}\selectfont ±1.4}  
&78.9\textsubscript{\fontsize{4pt}{4pt}\selectfont ±1.7} 
&\cellcolor{pink!50}81.5\textsubscript{\fontsize{4pt}{4pt}\selectfont ±2.4}

\\

\rowcolor{gray!20} Pretrained GDT-Diff   & \Checkmark

&\cellcolor{pink!50}73.2\textsubscript{\fontsize{4pt}{4pt}\selectfont ±1.3}  
&\cellcolor{pink!50}73.9\textsubscript{\fontsize{4pt}{4pt}\selectfont ±1.8} 
&73.5\textsubscript{\fontsize{4pt}{4pt}\selectfont ±2.1}  

&\cellcolor{pink!50}75.9\textsubscript{\fontsize{4pt}{4pt}\selectfont ±1.6}  
&\cellcolor{pink!50}78.4\textsubscript{\fontsize{4pt}{4pt}\selectfont ±1.4} 
&78.3\textsubscript{\fontsize{4pt}{4pt}\selectfont ±2.5} 

&83.7\textsubscript{\fontsize{4pt}{4pt}\selectfont ±1.8}  
&83.2\textsubscript{\fontsize{4pt}{4pt}\selectfont ±1.1}  
&85.9\textsubscript{\fontsize{4pt}{4pt}\selectfont ±2.0} 

&\cellcolor{pink!50}83.8\textsubscript{\fontsize{4pt}{4pt}\selectfont ±1.2}  
&85.1\textsubscript{\fontsize{4pt}{4pt}\selectfont ±1.3}  
&\cellcolor{pink!50}86.9\textsubscript{\fontsize{4pt}{4pt}\selectfont ±1.7}

&\cellcolor{pink!50}74.3\textsubscript{\fontsize{4pt}{4pt}\selectfont ±1.8}  
&77.0\textsubscript{\fontsize{4pt}{4pt}\selectfont ±1.5}  
&\cellcolor{pink!50}75.9\textsubscript{\fontsize{4pt}{4pt}\selectfont ±2.5} 

&76.2\textsubscript{\fontsize{4pt}{4pt}\selectfont ±1.9}  
&\cellcolor{pink!50}79.4\textsubscript{\fontsize{4pt}{4pt}\selectfont ±2.3} 
&81.0\textsubscript{\fontsize{4pt}{4pt}\selectfont ±1.6}

\\

\bottomrule

\end{tabular}
\end{table*}

Table \ref{comparison} presents a comprehensive comparison across six brain disorders on the Schaefer100 atlas \citep{schaefer2018local}. Traditional deep learning approaches (e.g., BrainGNN \citep{li2021braingnn} and BrainNetTF \citep{kan2022brain}) trained from scratch without pretraining. Classical graph diffusion–based methods, including Graph Diffusion Convolution (GDC) \citep{gasteiger2019diffusion}, Adaptive GDC \citep{zhao2021adaptive} and Graph Diffusion Transformer (GDT) \citep{wu2025supercharging}, improve robustness through predefined diffusion operators but remain constrained by fixed propagation schemes and the absence of task-agnostic pretraining.
Foundation models based on brain time-series data (BrainLM \citep{caro2023brainlm} and Brain-JEPA \citep{dong2024brain}) benefit from large-scale temporal pretraining but do not explicitly capture graph-structured connectome topology, while connectome-based foundation models (BrainMass \citep{yang2024brainmass}) further improve performance by leveraging connectivity patterns but are limited to static representations. In contrast, BrainGFM \citep{wei2025brain}, as a graph-based foundation model, explicitly integrates graph topology with pretraining objectives and consistently outperforms time-series– and connectome-based counterparts.
Notably, Pretrained GDT constitutes a minor yet novel contribution of this work, as we are the first to introduce a pretraining strategy for GDT. Among all variants, diffusion-based pretraining achieves the strongest overall performance, demonstrating that diffusion-aware objectives provide particularly informative supervision. However, applying diffusion-based pretraining to GDT results in only marginal gains compared to either pretrained GDT or diffusion-pretrained BrainGFM, which is expected since both the architecture and the pretraining objective encode global diffusion priors, leading to overlapping inductive biases and diminishing returns.

\begin{figure*}[ht]
\centering
	\includegraphics[width=17cm]{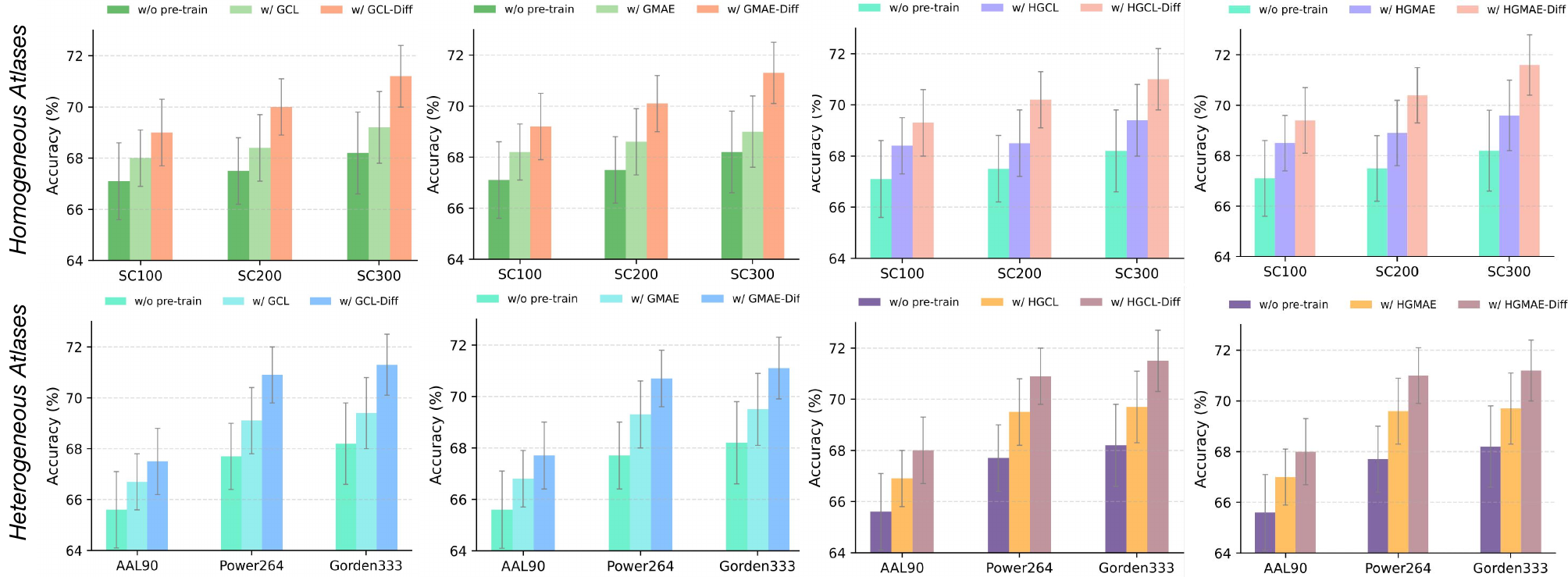}
    \vspace{-7pt}
	\caption{We evaluate graph- and hypergraph-based pretraining (denoted by G and H, respectively) with and without diffusion on homogeneous and heterogeneous atlases, where SC100/200/300 denote Schaefer atlases with 100/200/300 ROIs. Diffusion-based pretraining consistently improves performance across atlas settings on ABIDE.
 }
	\label{hoht}
\end{figure*}

\subsection{Effectiveness of Diffusion-based Pretraining across Graph and Hypergraph Paradigms}

As shown in Figure \ref{hoht}, this ablation study evaluates whether the proposed diffusion-based pretraining paradigm generalizes across different structural representations, including both graphs and hypergraphs, under homogeneous and heterogeneous atlas settings on the ABIDE dataset for ASD classification. We consider two representative pretraining paradigms: contrastive learning (GCL / HGCL) and masked autoencoder (GMAE / HGMAE), with and without diffusion-based objectives.
In the homogeneous atlas setting (top row in Figure \ref{hoht}), all models are constructed using Schaefer (SC) atlases with different spatial resolutions, where SC100, SC200, and SC300 denote Schaefer parcellations with 100, 200, and 300 ROIs, respectively. As the ROI resolution increases, diffusion-based pretraining consistently improves classification accuracy across both graph and hypergraph models, indicating that the proposed paradigm effectively captures global structural dependencies beyond resolution-specific node representations.
In the heterogeneous atlas setting (bottom row), models are evaluated across structurally diverse atlases, including AAL, Power, and Gordon, which differ substantially in parcellation strategies and ROI definitions. Despite this heterogeneity, diffusion-enhanced pretraining yields stable and consistent performance gains, demonstrating strong atlas-agnostic generalization.
Notably, the observed improvements hold for both graph-based (GCL, GMAE) and hypergraph-based (HGCL, HGMAE) pretraining paradigms.

\begin{figure*}[ht]
\centering
	\includegraphics[width=16.cm]{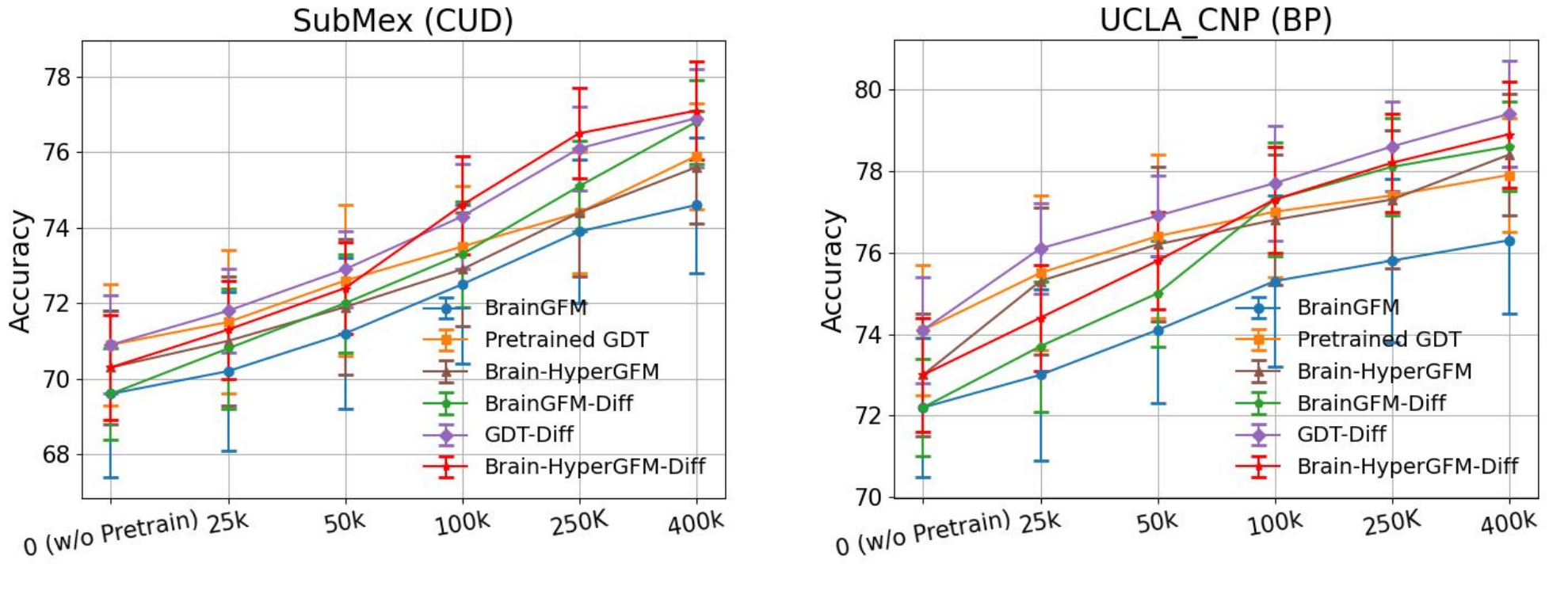}
    \vspace{-15pt}
	\caption{We compare diffusion-embedded architectures (GDT), diffusion-based pretraining (BrainGFM-Diff, Brain-HyperGFM), and their combination (GDT-Diff) on two datasets. Results highlight distinct behaviors of architectural diffusion and diffusion-based pretraining across data scales.
 }
	\label{data_size}
    \vspace{-10pt}
\end{figure*}

\subsection{Diffusion under Varying Pretraining Data Sizes}

As shown in Figure \ref{data_size}, the ablation study investigates how different strategies for incorporating diffusion mechanisms influence downstream disease classification performance as the amount of pretraining data increases. Specifically, we compare two orthogonal ways of leveraging diffusion:
(i) embedding diffusion directly into the model architecture (e.g., GDT), and
(ii) incorporating diffusion as a task-agnostic pretraining paradigm, as proposed in our diffusion-based graph and hypergraph foundation models.

We observe that diffusion-enhanced GDT and our diffusion-based pretrained models achieve comparable performance when sufficient pretraining data is available, suggesting that both strategies are capable of capturing global structural aggregation induced by diffusion processes. Notably, GDTs exhibits stronger performance in the low-data regime, which we attribute to its architectural diffusion bias that enables effective global information propagation even with limited pretraining.
In contrast, our diffusion-based pretraining paradigm shows superior scalability as the pretraining data size increases. Importantly, this performance gain is achieved without modifying the model architecture, highlighting that the improvement stems purely from a more effective pretraining rather than increased model capacity. This property is particularly desirable for foundation models, as it enables performance gains through data scaling while maintaining a fixed and reusable backbone.
Furthermore, pretraining diffusion-aware architectures such as GDT with our diffusion-based objective (GDT-Diff) yields only marginal improvements over their non-pretrained counterparts. This suggests that architectural diffusion mechanisms may already encode sufficient global aggregation priors, leading to diminishing returns when diffusion is applied again at the pretraining stage. These results indicate that, in practice, either architectural diffusion or diffusion-based pretraining is sufficient, and combining both does not necessarily lead to additive benefits.

\subsection{Efficiency and Computational Cost Analysis}


\begin{table}[h!]
\footnotesize
\centering
\caption{Efficiency comparison between GDT and BrainGFM variants.
“PT” denotes pretraining.}
\renewcommand{\arraystretch}{.9}
\setlength{\tabcolsep}{1pt}
\vspace{-8pt}
\begin{tabular}{c|ccc}
\midrule
Model & FLOPs (M) & Params (K) & Infer. Time (ms) \\ 
\midrule
\rowcolor{gray!10} GDT & 654.942 & 384.602 & 4.606 ± 1.004 \\ 
\rowcolor{gray!30} BrainGFM & \textbf{339.730}  & \textbf{287.710}  & \textbf{3.347 ± 0.703} \\
\rowcolor{gray!10} Brain-HyperGFM &342.456 &291.763 & 3.624 ± 0.825 \\ \midrule
\rowcolor{gray!10} GDT-PT & 654.942 & 384.602 & 2.254 ± 0.673 \\ 
\rowcolor{gray!30} BrainGFM-PT & \textbf{339.730}  & \textbf{287.710}  & \textbf{1.847 ± 0.614} \\ 
\rowcolor{gray!10} Brain-HyperGFM-PT &342.456 &291.763  &1.963 ± 0.633 \\
\bottomrule
\end{tabular}
\label{tab:flops_params_comparison}
\end{table}

Table \ref{tab:flops_params_comparison} reports the computational efficiency of different graph foundation models, including Graph Diffusion Transformer (GDT) and our proposed BrainGFM variants. GDT explicitly embeds graph diffusion operators into the Transformer architecture, leading to substantially higher computational overhead. In contrast, BrainGFM decouples diffusion from model inference by leveraging diffusion-aware pretraining, enabling more efficient downstream execution.

As shown in the table, BrainGFM reduces FLOPs by nearly 48\% and parameters by 25\% compared to GDT, while also achieving faster inference time. The hypergraph extension (Brain-HyperGFM) introduces only a marginal increase in computational cost, indicating that hypergraph modeling can be incorporated with minimal efficiency loss.
When pretraining is applied, all models benefit from faster inference, but BrainGFM-PT consistently remains the most efficient, achieving the lowest FLOPs, parameters, and inference latency. These results demonstrate that diffusion-aware pretraining provides a more computationally economical alternative to embedding diffusion (GDT) directly into the model architecture, making BrainGFM particularly suitable for large-scale and resource-constrained applications.

\section{Conclusion}

We propose a diffusion-based pretraining paradigm for brain graphs and hypergraphs that better aligns graph contrastive learning and graph masked autoencoder pretraining with the structural properties of brain connectomes.
By leveraging diffusion to guide graph and hypergraph dropping and masking for the augmentation, the proposed approach alleviates the limitations of random perturbations that either overly disrupt semantic connectivity patterns or provide insufficient diversity for effective pretraining.
In addition, diffusion is incorporated into graph-level readout and node-level masked reconstruction, enabling both contrastive views and reconstructed graphs to capture global structural information.

\section*{Impact Statement}

This work introduces a diffusion-based pretraining paradigm for both graph-level contrastive learning and node-level masked autoencoder pretraining on graphs and hypergraphs, providing a general framework for incorporating global structural information into graph representation learning.
The proposed paradigm is broadly applicable to other domains where global dependencies play a critical role, such as large-scale social networks, molecular graphs, and knowledge graphs.
The diffusion-guided augmentation strategy offers a general and adaptive alternative to random graph perturbations and can be readily applied to a wide range of graph augmentation scenarios.
In addition, the proposed diffusion-based readout can be seamlessly integrated into existing graph neural networks to generate high-quality graph embeddings, and the diffusion-based reconstruction mechanism naturally extends to generative graph models.
These components suggest a versatile and extensible paradigm with potential impact beyond neuroscience.

\nocite{langley00}

\bibliography{example_paper}
\bibliographystyle{icml2026}

\newpage
\appendix
\onecolumn

\section{Motivation}
The necessity of introducing graph diffusion into brain graph pretraining is motivated by the following key observations:
\begin{itemize}
    \item \textbf{Brain graphs are globally organized rather than locally defined.}  
    The functional role of a brain region is determined by its participation in distributed networks spanning multiple regions, rather than by immediate neighbors alone. Purely local message passing and neighborhood-based reconstruction fail to capture such long-range functional dependencies.

    \item \textbf{Random augmentation and masking are semantically unsafe for brain graphs.}  
    Naive random dropping or masking, commonly adopted in graph contrastive learning and masked autoencoders, can easily disrupt critical connectivity patterns or produce unrealistic contrastive views, leading to unstable training signals and fragile representations.

    \item \textbf{Existing graph-level readout schemes ignore global topology.}  
    Conventional pooling or attention-based readout functions treat nodes as an unordered set and aggregate representations without explicitly considering structural relationships, limiting their ability to encode brain-wide organization.

    \item \textbf{Local reconstruction in masked autoencoders suffers from locality bias.}  
    Most graph masked autoencoder methods reconstruct masked nodes or edges using only local neighborhood information, which is insufficient for brain graphs where distant but structurally related regions provide essential contextual cues.

    \item \textbf{Graph diffusion provides a principled inductive bias aligned with brain connectivity.}  
    By modeling multi-hop propagation and global structural influence, diffusion enables structure-aware augmentation, topology-aware readout, and global reconstruction, offering a unified solution to the above limitations.
\end{itemize}

\section{Comparison of different Graph Diffusion Kernels}

\begin{figure*}[htbp]
\centering
	\includegraphics[width=17cm]{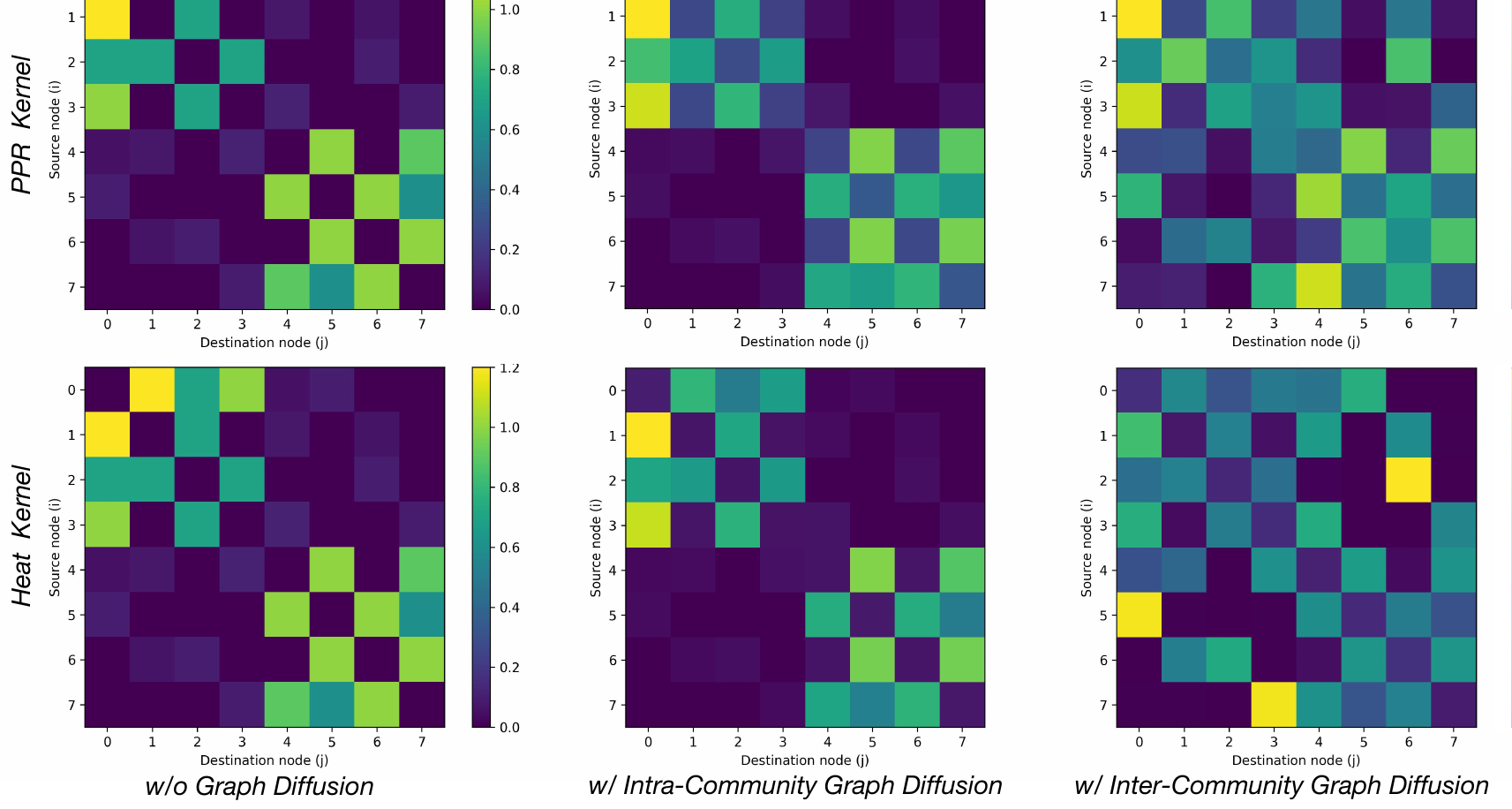}
	\caption{Each heatmap shows diffusion strengths between node pairs, where brighter colors indicate stronger connections. The first column depicts original intra-community connectivity, the second column applies diffusion within each community, and the third column applies diffusion across communities, illustrating both intra- and inter-community propagation effects. 
 }
	\label{commu}
\end{figure*}

As shown in Figure \ref{commu}, we provide a qualitative analysis of the diffusion behaviors of the Personalized PageRank (PPR) and Heat kernels. We construct a toy graph consisting of two disjoint communities, each containing four nodes, where initial connections only exist between neighboring nodes within the same community. The diffusion strength between node pairs is visualized as heatmaps, with brighter colors indicating stronger propagated connectivity.

In the first column, without graph diffusion, node interactions remain strictly localized. Each node is only connected to its immediate neighbors within the same community, reflecting the limited receptive field of direct adjacency-based modeling and the absence of long-range information propagation.

When diffusion is applied within individual communities (second column), both PPR and Heat kernels enable information to propagate beyond immediate neighbors. Nodes gradually establish connections with non-adjacent nodes inside the same community, demonstrating that diffusion effectively captures higher-order intra-community dependencies. Compared to direct adjacency, the resulting connectivity becomes smoother and more expressive, especially under the Heat kernel, which exhibits a more globally distributed propagation pattern.

More importantly, when diffusion is further extended across communities (third column), non-zero connectivity emerges between nodes belonging to different communities. This is evidenced by the appearance of diffusion mass in the off-diagonal blocks of the heatmaps, indicating successful inter-community information propagation. While PPR preserves a certain degree of locality through its restart mechanism, the Heat kernel promotes more uniform global diffusion, leading to stronger cross-community interactions.

Overall, although PPR and Heat kernels exhibit distinct propagation characteristics, both effectively expand the receptive field from local neighborhoods to global graph structures. This experiment confirms that graph diffusion enables global information aggregation beyond isolated subgraphs, providing a crucial foundation for the robustness and scalability of diffusion-based pretraining.

\begin{table}[htbp]
\centering
\footnotesize
\caption{Comparison of different graph diffusion kernels.}
\setlength{\tabcolsep}{6pt}
\renewcommand{\arraystretch}{1.15}
\begin{tabular}{l|ccc}
\hline
\textbf{Property} 
& \textbf{RW} 
& \textbf{PPR} 
& \textbf{Heat} \\
\hline

Formulation 
& $P^{k}$ 
& $\alpha \sum_{m=0}^{\infty}(1-\alpha)^m P^m$ 
& $e^{-tL}$ \\

Diffusion type 
& Discrete-step 
& Weighted multi-step 
& Continuous-time \\

Propagation range 
& Fixed ($k$-hop) 
& Adaptive (local $\leftrightarrow$ global) 
& Global \\

Path aggregation 
& Single-length paths 
& All lengths with decay 
& All lengths (spectral) \\

Local structure preservation 
& \ding{51} 
& \ding{51}\ding{51} 
& \ding{51} \\

Global information capture 
& \ding{55} 
& \ding{51} 
& \ding{51}\ding{51} \\

Robustness to node/edge drop 
& \ding{55} 
& \ding{51} 
& \ding{51}\ding{51} \\

Sensitivity to hyperparameters 
& High ($k$) 
& Moderate ($\alpha$) 
& Low ($t$) \\

\bottomrule
\end{tabular}
\end{table}

\begin{table}[htbp]
\footnotesize
\centering
\caption{Ablation study on different diffusion kernels on OASIS3 for DM Diagnosis.}
\renewcommand{\arraystretch}{1.}
\setlength{\tabcolsep}{4pt}
\begin{tabular}{c|ccc}
\midrule
Diffusion Kernels  &PPR & Heat & RW \\ 
\midrule
\rowcolor{gray!30} BrainGFM-Diff & 77.6\textsubscript{\fontsize{4pt}{4pt}\selectfont ±1.6}  & 77.3\textsubscript{\fontsize{4pt}{4pt}\selectfont ±1.8}  & 76.9\textsubscript{\fontsize{4pt}{4pt}\selectfont ±2.0}  \\ 
\rowcolor{gray!10} Brain-HyperGFM-Diff &77.8\textsubscript{\fontsize{4pt}{4pt}\selectfont ±1.4} &77.4\textsubscript{\fontsize{4pt}{4pt}\selectfont ±1.7}  &77.0\textsubscript{\fontsize{4pt}{4pt}\selectfont ±1.5} \\
\bottomrule
\end{tabular}
\label{tab:flops_params_comparison}
\end{table}

\section{Comparison among Graph, Hypergraph and Diffusion Hypergraph}

Tables~\ref{tab:width_vs_depth} and~\ref{tab:graph_hypergraph_diffusion} provide a unified view of graph modeling paradigms from the complementary perspectives of structural \emph{width} and propagation \emph{depth}.
Hypergraph modeling emphasizes width by explicitly encoding high-order relations through hyperedges, enabling parallel aggregation over node groups and naturally alleviating over-smoothing.
However, its information propagation is typically limited to single-hop group interactions and requires dedicated hypergraph operators with higher construction cost.
In contrast, graph diffusion focuses on depth, propagating information across multiple hops via analytical diffusion kernels, which enables global context integration while remaining compatible with standard GNN architectures.
Yet, diffusion alone operates on pairwise structures and does not explicitly model high-order relations.

The second table further highlights that graph diffusion and hypergraph modeling address orthogonal limitations of conventional graphs: diffusion extends structural depth, while hypergraphs expand relational width.
Their combination—diffusion hypergraphs—simultaneously supports high-order relations and global multi-hop propagation, yielding group-aware global representations with higher expressiveness.
This width–depth unification is particularly appealing for brain network modeling, where functional organization involves both higher-order interactions among regions and long-range dependencies across the entire network.

\begin{table*}[htbp]
\footnotesize
\centering
\renewcommand\arraystretch{1.25}
\setlength{\tabcolsep}{7pt}
\caption{Comparison between Hypergraph modeling and Graph Diffusion from the perspective of \textbf{Width} and \textbf{Depth}.}
\label{tab:width_vs_depth}
\begin{adjustbox}{max width=\linewidth}
\begin{tabular}{l|c|c}
\toprule
\textbf{Aspect} 
& \textbf{Hypergraph (Width)} 
& \textbf{Graph Diffusion (Depth)} \\
\midrule

\rowcolor{gray!10}
Core Perspective 
& \textbf{Structural Width} 
& \textbf{Propagation Depth} \\

Basic Modeling Unit 
& Hyperedge (multi-node relation) 
& Edge-based message passing \\

Long-range Dependency 
& \ding{51}\ Single-hop group connectivity 
& \ding{51}\ Multi-hop propagation \\

Information Flow Pattern 
& Parallel aggregation across node groups 
& Sequential propagation across layers \\

High-order Relation Modeling 
& \ding{51}\ Explicitly supported 
& \ding{55}\ Implicit / pairwise only \\

Structural Expressiveness 
& High (group-wise interactions) 
& Moderate (path-based interactions) \\

Risk of Over-smoothing 
& \ding{51}\ Naturally alleviated 
& \ding{55}\ Increases with depth \\

Compatibility with GNNs 
& \ding{55}\ Requires hypergraph operators 
& \ding{51}\ Plug-and-play with standard GNNs \\

Computational Cost 
& Higher (hyperedge construction) 
& Lower (standard graph operations) \\
\bottomrule
\end{tabular}
\end{adjustbox}
\end{table*}

\begin{table*}[htbp]
\footnotesize
\centering
\renewcommand\arraystretch{1.}
\setlength{\tabcolsep}{6pt}
\caption{Comparison of different graph modeling paradigms.}
\label{tab:graph_hypergraph_diffusion}
\begin{tabular}{l|c|c|c|c}
\toprule
\textbf{Aspect} 
& \textbf{Graph} 
& \textbf{Hypergraph} 
& \textbf{Diffusion Graph} 
& \textbf{Diffusion Hypergraph} \\
\midrule
Relation order 
& Pairwise 
& High-order 
& Global Multi-hop 
& High-order + Global Multi-hop \\

Structural extension 
& None 
& Width 
& Depth 
& Width $\times$ Depth \\

Explicit high-order modeling 
& \ding{55} 
& \ding{51} 
& \ding{55} 
& \ding{51} \\

Multi-hop propagation 
& \ding{55} 
& \ding{55} 
& \ding{51} 
& \ding{51} \\

Topology redefinition 
& \ding{55} 
& \ding{51} 
& \ding{55} 
& \ding{51} \\

Core representation 
& Adjacency $A$ 
& Incidence $H$ 
& Diffusion kernel $K_G$ 
& Hypergraph diffusion $K_{\mathcal{H}}$ \\

Modeling focus 
& Local connectivity 
& Group-wise relations 
& Global propagation 
& Group-aware global propagation \\

Expressiveness 
& Low 
& Medium 
& Medium 
& High \\
\bottomrule
\end{tabular}
\end{table*}

\section{Graph Dropping vs. Graph Masking}

\begin{table}[htbp]
\centering
\footnotesize
\caption{Comparison between graph augmentation via dropping and graph masking in graph pretraining.}
\renewcommand{\arraystretch}{1.15}
\setlength{\tabcolsep}{4pt}
\begin{tabular}{l|cc}
\midrule
\textbf{Aspect} & \textbf{Graph Augmentation (Drop)} & \textbf{Graph Masking (Mask)} \\
\midrule
Granularity
& Graph-level perturbation
& Node-level masking \\
Core operation
& Remove nodes, edges, or features
& Hide node or edge information \\
Information removal
& Irreversible (permanently discarded)
& Reversible (masked but retained) \\
Graph structure
& Graph topology is altered
& Graph topology is preserved \\
Learning objective
& View-invariant representation learning
& Context-aware reconstruction learning \\
Training paradigm
& Contrastive learning
& Masked autoencoder \\
Supervision signal
& Cross-view alignment
& Reconstruction loss \\
Requirement to recover
& Not required
& Explicitly required \\
Representation property
& Robustness and invariance
& Expressiveness and completeness \\
Typical methods
& GraphCL
& GraphMAE \\
Role in pretraining
& Encourage semantic consistency
& Encourage structural understanding \\
\bottomrule
\end{tabular}
\label{drop_mask}
\end{table}

\begin{table}[htbp]
\centering
\footnotesize
\caption{Comparison between random and diffusion-based drop/mask strategies for graph and hypergraph pretraining.}
\label{tab:drop_mask_comparison}
\setlength{\tabcolsep}{9pt}
\renewcommand{\arraystretch}{1.25}
\begin{tabular}{l|cc|cc}
\toprule
\textbf{Property} 
& \multicolumn{2}{c|}{\textbf{Random Strategies}} 
& \multicolumn{2}{c}{\textbf{Diffusion-based (Ours)}} \\
& Drop & Mask & Drop & Mask \\
\midrule

Structure-aware              
& \ding{55} & \ding{55} & \ding{51} & \ding{51} \\

Global information considered 
& \ding{55} & \ding{55} & \ding{51} & \ding{51} \\

Perturbation strength control 
& \ding{55} & \ding{55} & \ding{51} & \ding{51} \\

Moderate augmentation         
& \ding{55} & \ding{55} & \ding{51} & \ding{51} \\

Semantic safety               
& \ding{55} & \ding{55} & \ding{51} & \ding{51} \\

Preserve brain network backbone
& \ding{55} & \ding{55} & \ding{51} & \ding{51} \\

Contrastive view diversity    
& \ding{51} & \ding{51} & \ding{51} & \ding{51} \\

Training stability            
& \ding{55} & \ding{55} & \ding{51} & \ding{51} \\

Suitability for brain graphs  
& Low & Low & High & High \\
\bottomrule
\end{tabular}
\end{table}

Table \ref{drop_mask} highlights the fundamental differences between dropping-based graph augmentation and masking-based graph pretraining from the perspectives of granularity, supervision, and learning objectives. Dropping operates at the graph level, where stochastic perturbations are applied to construct multiple corrupted views of the same graph, and supervision is imposed by aligning their graph-level representations. Such a design encourages invariance to node or edge removal and improves robustness under structural perturbations, but does not require recovering the discarded information. In contrast, masking is typically formulated at the node level, where partial node or edge information is hidden while the overall topology is preserved, and the model is explicitly trained to reconstruct the masked components from contextual cues. Consequently, masking promotes fine-grained structural understanding and expressive representations through local reconstruction, whereas dropping emphasizes global semantic consistency across graph views. These two mechanisms therefore provide complementary and orthogonal supervisory signals—graph-level invariance versus node-level recoverability—making their combination particularly effective for learning generalizable graph representations.

\section{Comparison of Graph Readout}
\label{app:readout}

Conventional graph readout functions summarize a graph by aggregating node representations while treating the node set as unordered and largely structure-agnostic \citep{buterez2022graph}.
Typical choices include mean pooling and max pooling, which compute graph-level representations by simple statistical aggregation of node features,
\begin{equation}
\mathcal{R}{\mathrm{mean}} = \frac{1}{N}\sum{i=1}^{N} x_i,
\qquad
\mathcal{R}{\mathrm{max}} = \max{i=1,\dots,N} x_i,
\end{equation}
as well as attention-based pooling, which assigns adaptive importance weights to nodes via a learnable scoring function,
\begin{equation}
\alpha_i = \frac{\exp(w^\top x_i)}{\sum_{j=1}^{N} \exp(w^\top x_j)},
\qquad
\mathcal{R}{\mathrm{att}} = \sum{i=1}^{N} \alpha_i x_i,
\end{equation}
where $w\in\mathbb{R}^{d}$ is a trainable attention vector.
While attention-based readout introduces node-wise adaptivity, all these operators perform aggregation without explicitly incorporating the underlying graph or hypergraph topology at the readout stage.
As a result, the resulting graph-level representations may be sensitive to node or edge perturbations and exhibit instability across contrastive views, particularly in settings where global structural dependencies are critical.

Conventional graphs model only pairwise relationships between nodes, limiting their ability to capture complex interactions in structured data.
Hypergraphs extend this formulation by explicitly introducing hyperedges that connect multiple nodes, thereby enriching the relational representation along the \emph{width} dimension and enabling the modeling of high-order group-wise interactions.
In contrast, graph diffusion does not redefine the graph topology; instead, it enhances structural representations through multi-hop information propagation, enriching the modeling capacity along the \emph{depth} dimension by capturing long-range dependencies and global structural consistency.
Importantly, hypergraph modeling and diffusion-based modeling address two \emph{orthogonal} aspects of graph representation learning: the former focuses on \emph{what constitutes a relation} (relational width), while the latter focuses on \emph{how far information propagates} (structural depth).
Diffused hypergraphs naturally unify these two perspectives by enabling high-order group-aware relations to propagate over multiple hops, thereby jointly modeling collective interactions and long-range dependencies.
This unified formulation provides a more expressive and principled framework for representing complex systems, such as brain networks, where both functional modules and their global interactions are critical.

\begin{table}[htbp]

\renewcommand\arraystretch{.8}
\setlength{\tabcolsep}{10pt}
    \centering

    \caption{Performance improvements of different graph readout strategies on the ADNI dataset.
All reported gains are measured relative to the Traditional Readout baseline in terms of AUC, ACC, and F1.}
    \begin{tabular}{c|ccc}
        \toprule
       Methods & AUC  & ACC & F1  \\
        \midrule
          Traditional Readout &Baseline  &Baseline    &Baseline        \\
         Attention Readout  &1.3\%$\uparrow$ &1.5\%$\uparrow$  &1.9\%$\uparrow$  \\
         \midrule
         Diffusion Readout  &1.9\%$\uparrow$ &2.5\%$\uparrow$  &2.6\%$\uparrow$  \\
        \bottomrule
    \end{tabular}
        \label{bands}
\end{table}

\begin{table*}[htbp]
\footnotesize
\centering
\renewcommand{\arraystretch}{1.05}
\setlength{\tabcolsep}{9pt}
\caption{Comparison between conventional graph readout and our proposed diffusion-based graph readout.}
\label{tab:readout_comparison_check}
\begin{tabular}{l|c|c}
\toprule
\textbf{Dimension} 
& \textbf{Conventional Readout} 
& \textbf{Proposed Graph Diffusion Readout} \\
\midrule
Structural awareness 
& \ding{55} No 
& \ding{51} Yes \\

Node-wise information propagation 
& \ding{55} Absent
& \ding{51} Present \\

Robustness to node/edge dropping 
& \ding{55} Highly sensitive
& \ding{51} Robust \\

Representation semantics 
& \ding{55} Local nodes aggregation 
& \ding{51} Global topological state \\

Stability in graph contrastive learning 
& \ding{55} Low 
& \ding{51} High \\

Brain network interpretability 
& \ding{55} ROI-level 
& \ding{51} Network / subnetwork-level \\

\bottomrule
\end{tabular}
\end{table*}

\begin{table}[htbp]
\small
\centering
\caption{Comparison of different graph readout strategies.}
\label{tab:readout_comparison}
\renewcommand{\arraystretch}{1.}
\setlength{\tabcolsep}{9pt}
\begin{tabular}{lccc}
\toprule
\textbf{Property} 
& \textbf{Traditional Readout} 
& \textbf{Attention Readout} 
& \textbf{Diffusion Readout} \\
\midrule
Graph structure used in readout       
& \ding{55} 
& \ding{55} 
& \ding{51} \\

Topology-aware aggregation            
& \ding{55} 
& \ding{55} 
& \ding{51} \\

Multi-hop information integration     
& \ding{55} 
& \ding{55} 
& \ding{51} \\

Global information propagation        
& \ding{55} 
& \ding{55} 
& \ding{51} \\

Permutation invariance                
& \ding{51} 
& \ding{51} 
& \ding{51} \\

Robustness to node/edge perturbations 
& Low 
& Medium 
& High \\

Node adaptive weighting          
& \ding{55} (None)
& \ding{51} (Local Node-wise)
& \ding{51} (Global Graph-wise) \\

\bottomrule
\end{tabular}
\end{table}

\begin{table*}[htbp]
\centering
\caption{Comparison between architectural diffusion and diffusion-based pretraining.}
\label{tab:diffusion_comparison}
\small
\setlength{\tabcolsep}{4pt}
\renewcommand{\arraystretch}{1.2}
\begin{tabular}{l|c|c}
\toprule
\textbf{Aspect} 
& \textbf{Architectural Diffusion (GDT)} 
& \textbf{Diffusion-based Pretraining (Ours)} \\
\midrule
Diffusion injection location
& In-model (architecture) 
& Out-model (pretraining) \\

Diffusion integration stage 
& Model architecture 
& Pretraining paradigm \\

Introduces additional parameters 
& \ding{51} 
& \ding{55} \\

Requires model modification 
& \ding{51} 
& \ding{55} \\

Scales with pretraining data 
& Limited 
& \ding{51} \\

Low-data performance 
& Strong 
& Moderate \\

High-data performance 
& Comparable 
& Superior \\

Model reusability 
& Task-specific 
& High \\

Compatibility with hypergraphs 
& Limited 
& \textbf{\ding{51}} \\

\bottomrule
\end{tabular}
\end{table*}

\section{Latent Diffusion vs. Graph Diffusion}

Table~\ref{tab:latent_vs_graph_diffusion} highlights the fundamental differences between latent diffusion on graphs and our proposed graph diffusion paradigm.
Latent diffusion models operate in the learned embedding space and rely on iterative neural denoising to generate or refine node- or graph-level representations.
As a result, their interaction with graph topology is indirect and mediated by the encoder, making the diffusion dynamics sensitive to latent noise schedules and training stability.
In contrast, graph diffusion is performed directly on the graph or hypergraph structure using analytical kernels, enabling explicit multi-hop information propagation over nodes, edges, and hyperedges without introducing additional learnable parameters.
This structural diffusion naturally preserves global connectivity patterns and remains robust under node or edge dropping and masking.
Moreover, while latent diffusion is primarily designed for generative modeling, graph diffusion serves as a topology-aware representation propagation mechanism that is particularly well suited for self-supervised pretraining.
These properties make graph diffusion better aligned with brain graphs, where functional semantics are encoded in global connectivity rather than latent generative trajectories.

\begin{table}[htbp]
\centering
\footnotesize
\renewcommand{\arraystretch}{1.25}
\setlength{\tabcolsep}{6pt}
\caption{Comparison between latent diffusion on graphs and our  graph diffusion paradigm.}
\label{tab:latent_vs_graph_diffusion}
\begin{tabular}{l|c|c}
\toprule
\textbf{Dimension} 
& \textbf{Latent Diffusion on Graphs} 
& \textbf{Graph Diffusion (Ours)} \\
\midrule

Diffusion space
& Latent embedding space
& Graph / hypergraph structure \\

Diffusion object
& Node or graph representations
& Nodes, edges, and hyperedges \\

Core purpose
& Generative modeling
& Representation propagation \& pretraining \\

Relation to graph topology
& Indirect (via encoder)
& Explicitly topology-aware \\

Diffusion operator
& Learned neural denoiser
& Analytical diffusion kernel (RW / PPR / Heat) \\

Training requirement
& Requires iterative denoising
& Parameter-free diffusion \\

Stability under perturbation
& Sensitive to latent noise schedule
& Robust to node/edge drop and mask \\

Global context modeling
& Implicit
& Explicit multi-hop propagation \\

Suitability for brain graphs
& Limited
& Well-aligned with functional connectivity \\

\bottomrule
\end{tabular}
\end{table}

\newpage

\section{Algorithm Pipelines}

\begin{algorithm}[htbp]
\caption{Diffusion-Guided Graph / Hypergraph Contrastive Pretraining (GCL / HGCL)}
\label{alg:gcl_hgcl}
\textbf{Input}: Node features $X$, structure $\mathcal{S}$ (graph $A$ or hypergraph $I$), batch $\mathcal{B}$ \\
\textbf{Parameter}: Encoder $f_{\theta}$, diffusion depth $T$, coefficients $\{\beta_k\}$, temperature $\tau$ \\
\textbf{Output}: Contrastive loss $\mathcal{L}_{\mathrm{GCL}}$
\begin{algorithmic}[1]
    \STATE Construct transition matrix $P(\mathcal{S})$
    \STATE Compute diffusion kernel $K$

    \STATE Perform feature diffusion to capture global context:
    \STATE \hspace{1em} $X^{\mathrm{diff}} = KX$ \hfill \textit{(global information propagation)}

    \STATE Compute diffusion-based importance scores:
    \STATE \hspace{1em} Node score $s^{(v)}_i = \|(X^{\mathrm{diff}})_i\|_2$
    \STATE \hspace{1em} Structure score $s^{(e)}$ or $s^{(h)}$ from $K$

    \STATE Compute \textbf{moderate diffusion-guided drop probabilities}:
    \STATE \hspace{1em} $p^{\mathrm{drop}} = \Psi(s)$

    \STATE Sample diffusion-guided drop masks:
    \STATE \hspace{1em} $\mathcal{M}^{(1)}, \mathcal{M}^{(2)} \sim \mathrm{Bernoulli}(1 - p^{\mathrm{drop}})$

    \STATE Generate two \textbf{diffusion-consistent augmented views}:
    \STATE \hspace{1em} $(X^{(1)}, \mathcal{S}^{(1)}) = \mathrm{Drop}(X^{\mathrm{diff}}, \mathcal{S}, \mathcal{M}^{(1)})$
    \STATE \hspace{1em} $(X^{(2)}, \mathcal{S}^{(2)}) = \mathrm{Drop}(X^{\mathrm{diff}}, \mathcal{S}, \mathcal{M}^{(2)})$

    \STATE Encode both views:
    \STATE \hspace{1em} $Z^{(1)} = f_{\theta}(X^{(1)}, \mathcal{S}^{(1)})$
    \STATE \hspace{1em} $Z^{(2)} = f_{\theta}(X^{(2)}, \mathcal{S}^{(2)})$

    \STATE \textbf{Graph diffusion readout} to obtain graph-level embeddings:
    \STATE \hspace{1em} $g^{(1)} = \mathcal{R}_{\mathrm{diff}}(Z^{(1)}, K)$
    \STATE \hspace{1em} $g^{(2)} = \mathcal{R}_{\mathrm{diff}}(Z^{(2)}, K)$

    \STATE Compute NT-Xent contrastive loss:
    \STATE \hspace{1em} $\mathcal{L}_{\mathrm{GCL}} = \mathrm{NTXent}(g^{(1)}, g^{(2)}, \tau)$

    \STATE \textbf{return} $\mathcal{L}_{\mathrm{GCL}}$
\end{algorithmic}
\end{algorithm}

\begin{algorithm}[htbp]
\caption{Graph / Hypergraph Diffusion Readout}
\label{alg:diff_readout}
\textbf{Input}: Node embeddings $Z \in \mathbb{R}^{N\times d}$, structure $\mathcal{S}$ (graph $A$ or hypergraph $I$) \\
\textbf{Parameter}: Diffusion kernel $K$ \\
\textbf{Output}: Graph-level embedding $g$
\begin{algorithmic}[1]
    \STATE Construct diffusion kernel from structure:
    \STATE \hspace{1em} $K = \mathcal{K}(\mathcal{S})$ 
    \hfill \textit{(graph or hypergraph diffusion)}

    \STATE Diffuse node embeddings to aggregate global context:
    \STATE \hspace{1em} $\tilde{Z} = KZ$

    \STATE Aggregate diffused embeddings using permutation-invariant pooling:
    \STATE \hspace{1em} $g = \mathrm{Pool}(\tilde{Z})$

    \STATE \textbf{return} $g$
\end{algorithmic}
\end{algorithm}

\begin{algorithm}[htbp]
\caption{Diffusion-Enhanced Graph / Hypergraph Masked Autoencoder (GMAE / HGMAE)}
\label{alg:gmae_hgmae}
\textbf{Input}: Node features $X$, structure $\mathcal{S}$ (graph $A$ or hypergraph $I$) \\
\textbf{Parameter}: Encoder $f_{\theta}$, decoder $g_{\phi}$, diffusion depth $T$, coefficients $\{\beta_k\}$, mask ratio $\rho$ \\
\textbf{Output}: Reconstruction loss $\mathcal{L}_{\mathrm{GMAE}}^{\mathrm{diff}}$
\begin{algorithmic}[1]
    \STATE Construct transition matrix $P(\mathcal{S})$
    \STATE Compute diffusion kernel $K$
    \STATE Perform feature diffusion:
    \STATE \hspace{1em} $X^{\mathrm{diff}} = KX$ \hfill \textit{(global information propagation)}

    \STATE Compute diffusion-based node importance scores:
    \STATE \hspace{1em} $s^{(v)}_i = \|(X^{\mathrm{diff}})_i\|_2$

    \STATE Compute \textbf{moderate diffusion-guided mask probabilities}:
    \STATE \hspace{1em} $p^{\mathrm{mask}} = \Phi(s)$

    \STATE Sample diffusion-guided masked node set $\mathcal{M}$ with ratio $\rho$
    \STATE Construct masked input $\tilde{X}$

    \STATE Encode diffusion-informed representations:
    \STATE \hspace{1em} $Z = f_{\theta}(\tilde{X}, \mathcal{S})$

    \STATE \textbf{Diffusion-supported reconstruction}:
    \STATE \hspace{1em} $\hat{X}^{\mathrm{diff}} = K Z$ \hfill \textit{(aggregate global context)}

    \STATE Reconstruct masked node features:
    \STATE \hspace{1em} $\hat{X}_{\mathcal{M}} = g_{\phi}(\hat{X}^{\mathrm{diff}}_{\mathcal{M}})$

    \STATE Compute node reconstruction loss:
    \STATE \hspace{1em} $\mathcal{L}_{\mathrm{node}} = \frac{1}{|\mathcal{M}|}\sum_{i\in\mathcal{M}}\|X_i - \hat{X}_i\|_2^2$

    \STATE \IF{structure reconstruction is enabled}
        \STATE Compute diffusion-supported connectivity target:
        \STATE \hspace{1em} $\widetilde{\mathcal{S}} = \mathcal{D}(\mathcal{S}, K)$
        \STATE Predict reconstructed structure $\widehat{\widetilde{\mathcal{S}}}$
        \STATE Compute structure reconstruction loss $\mathcal{L}_{\mathrm{struct}}$
    \ENDIF

    \STATE $\mathcal{L}_{\mathrm{GMAE}}^{\mathrm{diff}} = \mathcal{L}_{\mathrm{node}} + \eta\,\mathcal{L}_{\mathrm{struct}}$
    \STATE \textbf{return} $\mathcal{L}_{\mathrm{GMAE}}^{\mathrm{diff}}$
\end{algorithmic}
\end{algorithm}

\end{document}